\PassOptionsToPackage{table,dvipsnames}{xcolor}
\pdfoutput=1

\documentclass[11pt]{article}

\usepackage[final]{acl}

\usepackage{times}
\usepackage{latexsym}

\usepackage[T1]{fontenc}

\usepackage[utf8]{inputenc}

\usepackage{microtype}

\usepackage{inconsolata}

\usepackage{graphicx}
\usepackage{multirow}
\usepackage{booktabs}
\usepackage{caption}
\usepackage{lipsum}
\usepackage[dvipsnames]{xcolor}
\usepackage{xspace}
\usepackage{CJKutf8}
\usepackage[utf8]{inputenc}
\usepackage{amsfonts,amsmath,amssymb,amsthm}
\usepackage{bbm}
\usepackage{url}
\usepackage{titletoc}
\usepackage{fancyhdr}
\usepackage{enumitem}
\usepackage{subcaption}
\usepackage{nicefrac}
\usepackage{wrapfig}
\usepackage{tcolorbox}

\title{How Instruction and Reasoning Data shape Post-Training: \\Data Quality through the Lens of Layer-wise Gradients}

\author{Ming Li\textsuperscript{1}, 
Yanhong Li\textsuperscript{2}, 
Ziyue Li\textsuperscript{1}, 
Tianyi Zhou\textsuperscript{1} \\
  \textsuperscript{1}University of Maryland; 
  \textsuperscript{2}Allen Institute for AI\\
  \texttt{\{minglii, tianyi\}@umd.edu} \\
  Project: \url{https://github.com/MingLiiii/Gradient_Unified}
  }

\begin{document}

\maketitle

\begin{abstract}
As the post-training of large language models (LLMs) advances from instruction-following to complex reasoning tasks, understanding how different data affect finetuning dynamics remains largely unexplored. 
In this paper, we present a spectral analysis of layer-wise gradients induced by low/high-quality instruction and reasoning data for LLM post-training. 
Our analysis reveals that widely-studied metrics for data evaluation, e.g., IFD, InsTag, Difficulty, and Reward, can be explained and unified by spectral properties computed from gradients' singular value decomposition (SVD). Specifically, higher-quality data are usually associated with lower nuclear norms and higher effective ranks. 
Notably, effective rank exhibits better robustness and resolution than nuclear norm in capturing subtle quality differences. For example, reasoning data achieves substantially higher effective ranks than instruction data, implying richer gradient structures on more complex tasks.
Our experiments also highlight that models within the same family share similar gradient patterns regardless of their sizes, whereas different model families diverge significantly. 
Providing a unified view on the effects of data quality across instruction and reasoning data, 
this work illuminates the interplay between data quality and training stability, shedding novel insights into developing better data exploration strategies for post-training. 
\end{abstract}


\section{Introduction}

Large language models (LLMs) have shown remarkable potential for various complex tasks~\citep{zhao2023survey,xu2024survey,wu2025usablexai10strategies}, yet their success in real-world applications hinges not just on model size but also on the quality of data used for training~\citep{NEURIPS2020_1457c0d6, wang2023aligning, zhou2023lima}.
It is widely recognized that high-quality, diverse, and complex instruction-following examples during post-training~\citep{xu2023wizardlm,ding2023enhancing, Li2023ReflectionTuningDR,liu2023makes, wu2024laminilm, li2024mosaicitfreecompositionaldata,li2025rulerimprovingllmcontrollability}, such as supervised fine-tuning (SFT), is crucial for eliciting generalization performance and reliability.
There is growing recent interest in automated metrics that can evaluate data quality and select data for more efficient and effective post-training 
~\citep{chen2023alpagasus, cherry, Li2024SuperfilteringWD}. 
Moreover, beyond instruction-following, the reasoning capability of LLMs~\citep{openai2024openaio1card, deepseekai2025deepseekr1incentivizingreasoningcapability} has also been proven to be largely dependent on data quality.
For instance, s1.1~\citep{muennighoff2025s1simpletesttimescaling} utilizes only $1$k difficult math problems and DeepSeek-R1 generated responses to elicit LLM's strong reasoning capability. 

Despite the verified importance of data quality to post-training, \textbf{how the quality of instruction/reasoning data 
affect the gradients during post-training} still remains largely unexplored. In addition, \textbf{can we \textit{unify} different data quality metrics?} 
Prior work has mostly treated data quality filtering as a preprocessing step, evaluating its benefits in terms of end-task performance. But a systematic study is lacking to reveal the mechanism of how data quality affects the training dynamics. 
Meanwhile, there does not exist work that compares the learning dynamics induced by reasoning data and general instruction-following data or compares different data quality metrics' effects on post-training with data selection.  
A recent study of LLM post-training~\citep{li2024happenedllmslayerstrained} on fast vs. slow thinking~\citep{kahneman2011thinking} for the first time analyzes the training dynamics on different data through the lens of layer-wise gradients. It discovers that learning detailed intermediate reasoning steps leads to smaller and more stable gradient updates than learning final answers only. 
Yet, this study focuses only on comparing ``fast thinking'' (a few answer tokens) vs. ``slow thinking'' (CoT paths) but does not extend to more challenging problems requiring more complex reasoning. The metric used to measure gradients is limited to magnitude instead of more sophisticated spectral properties. 
\looseness-1

We address these gaps by conducting \textbf{a layer-wise, gradient-based spectral analysis} of LLM post-training when using \textbf{instruction/reasoning data of low/high quality.} 
Our study spans multiple diverse LLM families, including Qwen2.5 ~\citep{qwen2025qwen25technicalreport}, Llama3.1, Llama3.2~\citep{dubey2024llama3herdmodels}, Gemma2~\citep{gemmateam2024gemma2improvingopen}), and different sizes ($1.5$B - $14$B), to ensure the generality of our findings.
Inspired by s1~\citep{muennighoff2025s1simpletesttimescaling}, which selects data by difficulty, for reasoning data, we compare the s1.1 data and GSM8K~\citep{cobbe2021trainingverifierssolvemath} data (response generated by DeepSeek-R1) as low/high-quality data, respectively.  
For general instruction-following data, we adopt WizardLM~\citep{xu2023wizardlm}, Magpie~\citep{xu2024magpiealignmentdatasynthesis}, and OpenHermes 2.5~\citep{OpenHermes2.5} for our experiments and leverage automatic metrics for data quality, such as IFD~\citep{cherry}, InsTag~\citep{lu2023instag}, Difficulty, and Reward to partition these datasets into low/high-quality subsets. While the first three evaluate the instructions, the reward directly measures the response quality.  
\looseness-1

We developed several novel metrics measuring the spectral properties of gradients revealed by Singular Value Decomposition (SVD) (\textbf{SVD-based Metrics}) and Gradient \textbf{Similarity-based Metrics}, which are applied to the projection layers for Query, Key, Value, and Output in transformer architectures \citep{NIPS2017_3f5ee243}: 
(1) \textit{Nuclear Norm} measures the magnitude of the gradient, indicating the amount of changes and efforts required for post-training. 
(2) \textit{Effective Rank} \citep{roy2007effective} captures the dimensionality of the gradient, indicating the diversity of the gradient directions.
(3) \textit{Same-layer Similarity} measures the alignment between gradients of different projections within the same layer. 
(4) \textit{Adjacent-layer Similarity} measures the alignment of gradients between consecutive layers. 
\looseness-1

\textbf{\textit{Main Contributions:}}
In this study, we present a spectral analysis of layer-wise gradients in modern LLMs when finetuned on datasets of varying quality, namely high-quality, low-quality, instruction-following, and reasoning data. 
For the broad applicability of our conclusions, we conduct empirical investigations across diverse pretrained LLMs from multiple model families and on different datasets. 
Various automatic data evaluation metrics are used to split the data into low/high-quality partitions, and advanced reasoning data are also included for comparison. 
\textbf{Notably, we are the first to unify the effects of different data quality metrics and compare the instruction vs. reasoning data through the lens of layer-wise gradients. }
Unlike existing work that focuses primarily on the gradient magnitude, we incorporate both SVD-based and similarity-based metrics, offering a more comprehensive analysis. 
These findings reveal previously overlooked gradient patterns and provide insights into enhancing the stability and efficiency of data synthesis and LLM training.
\looseness-1

\textbf{\textit{Our key findings:}}
\begin{enumerate}[label=\textbf{\arabic*.}, leftmargin=4mm]
    \vspace{-2mm}
    \item Existing data quality metrics, e.g., IFD, InsTag, Difficulty, and Reward, can be \textbf{\textit{unified}} due to consistent spectral properties of gradients, i.e., lower nuclear norms and higher effective ranks on high-quality data. This finding extends to both instruction and reasoning data, providing a unified view of the data quality effects. \looseness-1
    \vspace{-2mm}
    \item Effective rank outperforms nuclear norm to distinguish low- vs. high-quality data. For reasoning data, s1.1 data yields the largest effective ranks across all experiments, suggesting high correlations between reasoning complexity and gradient diversity.
    \vspace{-2mm}
    \item Within the same model family, layer-wise gradients' spectral properties remain consistent across different model sizes. In contrast, the gradient patterns diverge significantly across distinct model families, reflecting the unique learning dynamics of each model family.
    \vspace{-2mm}
    \item Cosine similarities between gradients from the same layer and adjacent layers remain nearly zero for different types of data, so they cannot reflect data quality.  
\end{enumerate}

\section{Methodology}

\subsection{Preliminaries}

We investigate gradient behaviors under the most widely adopted SFT approach. 
Each data point in an SFT dataset $D$ consists of a pair $(x, y)$, where $x$ is the instruction and $y$ is the corresponding response. 
For the reasoning data, $y$ concatenates both thinking tokens and response tokens. 
Let $p_\theta$ be a large language model with parameters~$\theta$. 
Under the SFT paradigm, $p_\theta$ is finetuned on each pair $(x, y)$ by minimizing the following loss, where $y_j$ denotes the $j$-th token of $y$, $y_{<j}$ denotes the preceding tokens, and $l$ is the total length of $y$:
\vspace{-1mm}
\begin{equation}
L_\theta 
\;=\;
\frac{1}{l} \sum_{j=1}^{l} -\log\!\bigl(p_\theta(y_j \mid x,\, y_{<j})\bigr).
\label{eq:sft_loss}
\end{equation}
\vspace{-3mm}

In this paper, we focus on the gradients of the layers associated with the attention mechanism \citep{NIPS2017_3f5ee243}, namely the \emph{Query (Q)}, \emph{Key (K)}, \emph{Value (V)} projection layers and the final \emph{Output (O)} projection layer. 
For simplicity, we denote the gradients of these projection layers by $G_{Q,i}$, $G_{K,i}$, $G_{V,i}$, and $G_{O,i}$ for each layer $i \in \{0,1,\ldots,N-1\}$ in the LLM. 

\subsection{Gradient Metrics from Spectral Analysis}

To quantitatively analyze these gradients, we employ two categories of metrics: 
\textbf{(1)}~\emph{SVD-based Metrics} (the \textbf{Nuclear Norm} and the \textbf{Effective Rank}) and 
\textbf{(2)}~\emph{Similarity-based Metrics} (the \textbf{Same-layer Similarity} and the \textbf{Adjacent-layer Similarity}). 
While the SVD-based metrics describe properties of an individual gradient matrix, the similarity-based metrics reveal how gradients compare across different projections or layers.

\subsubsection{SVD-based Metrics}

Consider a gradient matrix $G_{X,i} \in \mathbb{R}^{m \times n}$, where $X \in \{Q,K,V,O\}$ indicates the projection (Query, Key, Value, or Output), and $i$ represents the index the Transformer layer. 
We can write its Singular Value Decomposition as
\vspace{-2mm}
{
\[
G_{X,i}
\;=\;
U\,\Sigma\,V^{T},
\]}
\vspace{-2mm}
where $U \in \mathbb{R}^{m \times m}$ and $V \in \mathbb{R}^{n \times n}$ are orthogonal matrices, and $\Sigma \in \mathbb{R}^{m \times n}$ is a diagonal matrix containing the singular values $\sigma_{1}, \dots, \sigma_{\min(m,n)}$, sorted in decreasing order. For simplicity, we omit all the subscripts ($X$ and $i$) for SVD matrices and singular values. 

\paragraph{Nuclear Norm.}
We measure the overall magnitude of the gradient matrix $G_{X,i}$ by its nuclear norm $\mathcal{N}_{X,i}$, which is formulated as
\vspace{-2mm}
\begin{equation}
\label{eq:nuclear_norm}
\mathcal{N}_{X,i}
\;=\;
\|G_{X,i}\|_{*}
\;=\;
\sum_{j=1}^{\min(m,n)}
\sigma_{j}.
\end{equation}
\vspace{-2mm}
A higher nuclear norm indicates a larger overall gradient scale, implying that the model parameters at that layer are being updated more significantly, further indicating a potential distribution shift between the response and the model to be trained. 
\looseness-1


\paragraph{Effective Rank.}
We measure how uniformly the singular values of \(G_{X,i}\) are distributed by the effective rank \(\mathcal{R}_{X,i}\).
We normalize the singular values and formulate the effective rank as
\vspace{-2mm}
{
\begin{align*}
\mathcal{R}_{X,i}
&= \exp\Bigl(
  -\sum_{j=1}^{\min(m,n)}
  \tilde{\sigma}_{j}
  \ln(\tilde{\sigma}_{j})
\Bigr)
\\
\tilde{\sigma}_{j}
&= \frac{\sigma_{j}}{\sum_{k=1}^{\min(m,n)}\sigma_{k}}
\end{align*}
}
\vspace{-2mm}
If only a few singular values are large (i.e., the gradient is concentrated in just a few directions), the effective rank is small. If many singular values all contribute significantly, the effective rank is relatively larger.
It measures how diverse the directions of the gradient are. 
A higher effective rank indicates the gradient is spread out over more directions, suggesting richer updates, whereas a smaller value means that only a few directions dominate the gradient directions. 

\subsubsection{Similarity-based Metrics}

While the SVD-based metrics above characterize individual gradient matrices in isolation, it can be equally insightful to analyze how gradients relate across projections or layers. 
To this end, we introduce cosine similarity measures at two levels: within the same layer and across adjacent layers.

\section{Experimental Setup}

\subsection{Models}
We evaluate our method on several pretrained LLMs across multiple families, covering a range of parameter scales. Specifically, we use Qwen2.5 \citep{qwen2025qwen25technicalreport} in four configurations (1.5B, 3B, 7B, and 14B parameters), Llama3.1 \citep{dubey2024llama3herdmodels} with 8B parameters, Llama3.2 in 1B and 3B configurations, and Gemma2 \citep{gemmateam2024gemma2improvingopen} in 2B and 9B configurations. 

\subsection{Datasets}


\textbf{Instruction-following data: }
\textbf{WizardLM} \citep{xu2023wizardlm} is an instruction-following dataset created via an LLM-based ``evolution'' strategy that iteratively rewrites a set of initial prompts into more complex multi-step instructions, automatically generating high-complexity queries beyond what human annotators typically produce. 
\textbf{Magpie} \citep{xu2024magpiealignmentdatasynthesis} is a fully synthetic alignment dataset obtained by prompting an aligned language model, and we utilize the $300$k high-quality subset selected by the authors for our source data. 
\textbf{OpenHermes 2.5} is a large-scale curated compilation of roughly one million instruction–response samples drawn from a diverse range of open-source and GPT-4-generated datasets, designed to maximize diversity and task coverage for robust fine-tuning.



\textbf{Reasoning data:}
We employ two sources of reasoning data, each paired with step-by-step traces and final solutions generated by DeepSeek-R1 \citep{deepseekai2025deepseekr1incentivizingreasoningcapability}: \textbf{s1.1K} data is provided by s1, which contains difficult math problems with responses generated by DeepSeek-R1. This data can be viewed as high-quality reasoning data since it succeeds in eliciting LLMs' reasoning capability with only $1,000$ samples. 
To curate the relatively low-quality reasoning data, motivated by s1's success in utilizing difficulty as the metric, we utilize the relatively easy math problems from \textbf{GSM8K} \citep{cobbe2021trainingverifierssolvemath} with DeepSeek-R1 generated responses. 

\vspace{-2mm}
\subsection{Data Quality Metrics to Partition Low/High-Quality Data}


We adopt four automated data evaluation metrics to analyze instruction-following data quality. We focus on data evaluation metrics that do not rely on additional evaluation sets or training, which might lead to customized shifting, while we aim at the effects of original properties of the data instances. 

\textbf{IFD}~\citep{cherry} quantifies the instruction-following difficulty of a sample by computing the ratio between the model’s perplexity when predicting the response without the instruction and with the instruction. A higher IFD indicates that the model fails to benefit from the instruction, suggesting that the instruction is either ambiguous or unhelpful. Following~\citet{Li2024SuperfilteringWD}, we use a small GPT2 model to efficiently compute IFD scores over large datasets.

\textbf{InsTag}~\citep{lu2023instag} performs open-set multi-label tagging over instructions, capturing semantic attributes such as domain, task type, and intent. We use two derived metrics: (1) \textit{instruction complexity}, defined by the number of tags per sample, and (2) \textit{instruction diversity}, defined by tag vocabulary coverage across the dataset. Higher complexity scores typically correspond to more elaborate, multifaceted instructions. We use the per-sample complexity score as the filtering signal.

\textbf{Reward Model Score} uses \textit{sfairXC/FsfairX-LLaMA3-RM-v0.1}, a reward model fine-tuned via preference modeling to predict human-aligned helpfulness \citep{xiong2024iterative}. Given a data pair, the model outputs a reward score reflecting predicted alignment with human preferences. 

\textbf{GPT-4o Difficulty Rating} prompts the GPT-4o model to assign a difficulty score ($from $1$ to 10$) to each instruction based on the perceived complexity, ambiguity, and reasoning depth required. This method approximates a human-aligned evaluation of instruction difficulty. 

For each dataset described above, and for each metric, we select $200$ samples with the highest scores and $200$ with the lowest scores for calculating the gradients. This allows us to isolate what each metric considers ``good-quality'' and ``low-quality'' data and to conduct gradient-based analysis across these contrasting subsets.

\section{Empirical Analysis}

\subsection{Instruction-Following Data}

\subsubsection{Unifying Different Data Quality Metrics}

\begin{table*}[t]
\centering
\resizebox{\textwidth}{!}{%
\begin{tabular}{l|l|
                c c c c|
                c c c c|
                c c c c|
                c c c c}
\toprule
\multirow{2}{*}{\textbf{Dataset}} & \multirow{2}{*}{\textbf{Metrics}} 
& \multicolumn{8}{c|}{\textbf{Nuclear Norm}}
& \multicolumn{8}{c}{\textbf{Effective Rank}} \\
\cmidrule(lr){3-6}\cmidrule(lr){7-10}\cmidrule(lr){11-14}\cmidrule(lr){15-18}
 & & \textbf{Proj} & High & Low & Gap 
   & \textbf{Proj} & High & Low & Gap
   & \textbf{Proj} & High & Low & Gap
   & \textbf{Proj} & High & Low & Gap \\
\midrule
\multirow{8}{*}{\textbf{WizardLM}}
& \multirow{2}{*}{\textbf{IFD}}
  & k & 1.3 & 6.1 & \cellcolor{red!20}{-4.8}
  & q & 1.4 & 4.6 & \cellcolor{red!20}{-3.2}
  & k & 88.5 & 14.2 & \cellcolor{green!20}{74.3}
  & q & 131.9 & 12.7 & \cellcolor{green!20}{119.2} \\
& 
  & v & 2.5 & 10.9 & \cellcolor{red!20}{-8.4}
  & o & 2.7 & 8.3 & \cellcolor{red!20}{-5.6}
  & v & 110.9 & 13.6 & \cellcolor{green!20}{97.3}
  & o & 165.2 & 12.5 & \cellcolor{green!20}{152.7} \\
\cmidrule(lr){2-18}
& \multirow{2}{*}{\textbf{InsTag}}
  & k & 1.9 & 4.1 & \cellcolor{red!20}{-2.2}
  & q & 1.9 & 3.2 & \cellcolor{red!20}{-1.3}
  & k & 95.6 & 19.9 & \cellcolor{green!20}{75.7}
  & q & 141.2 & 21.5 & \cellcolor{green!20}{119.7} \\
& 
  & v & 3.1 & 7.6 & \cellcolor{red!20}{-4.5}
  & o & 3.3 & 5.9 & \cellcolor{red!20}{-2.6}
  & v & 120.7 & 20.8 & \cellcolor{green!20}{99.9}
  & o & 180.5 & 22.3 & \cellcolor{green!20}{158.2} \\
\cmidrule(lr){2-18}
& \multirow{2}{*}{\textbf{Difficulty}}
  & k & 1.9 & 3.5 & \cellcolor{red!20}{-1.6}
  & q & 1.9 & 2.7 & \cellcolor{red!20}{-0.8}
  & k & 91.5 & 18.5 & \cellcolor{green!20}{73.0}
  & q & 133.2 & 20.0 & \cellcolor{green!20}{113.2} \\
& 
  & v & 3.1 & 6.8 & \cellcolor{red!20}{-3.7}
  & o & 3.3 & 5.2 & \cellcolor{red!20}{-1.9}
  & v & 114.9 & 19.1 & \cellcolor{green!20}{95.8}
  & o & 167.8 & 20.7 & \cellcolor{green!20}{147.1} \\
\cmidrule(lr){2-18}
& \multirow{2}{*}{\textbf{Reward}}
  & k & 1.2 & 4.3 & \cellcolor{red!20}{-3.1}
  & q & 1.2 & 3.6 & \cellcolor{red!20}{-2.4}
  & k & 91.5 & 35.6 & \cellcolor{green!20}{55.9}
  & q & 131.4 & 38.6 & \cellcolor{green!20}{92.8} \\
&
  & v & 2.1 & 7.8 & \cellcolor{red!20}{-5.7}
  & o & 2.3 & 6.5 & \cellcolor{red!20}{-4.2}
  & v & 113.2 & 36.9 & \cellcolor{green!20}{76.3}
  & o & 166.8 & 41.4 & \cellcolor{green!20}{125.4} \\
\bottomrule
\end{tabular}
}
\vspace{-2mm}
\caption{Nuclear norms and effective ranks of gradients calculated from high- or low-quality data selected by different metrics. \textit{High} represents the high-quality subset, while \textit{Low} represents the low-quality subset. \textit{Gap} is calculated by \textit{High} $-$ \textit{Low}. 
\textbf{All data quality metrics consistently identify the high-quality data, which show similar spectral properties of gradients: lower nuclear norms and higher effective ranks. Hence, the gradient properties can unify all metrics.} }
\vspace{-4mm}
\label{tab:411}
\end{table*}

In this section, we compare the effects of different data filtering metrics for general instruction-following data toward the gradient properties, including the averaged nuclear norms and effective ranks, as shown in Table \ref{tab:411}.
The nuclear norms in the table are calculated across the layers, $\mathcal{N}_{X} = \sum_{i=0}^{N-1}\mathcal{N}_{X,i}$, and the effective ranks are similar. 
Specifically, we use the WizardLM~\citep{xu2023wizardlm} data as the source data and calculate gradients on the Qwen2.5-7B model, and the data filtering metrics include IFD~\citep{cherry} (calculated on the Qwen2.5-7B model), InsTag~\citep{lu2023instag}, Difficulty, and Reward. 

As shown in the table, across all metrics, the high-quality subsets exhibit substantially smaller averaged nuclear norms in the layer-wise gradients, represented by the consistent negative values (in red) for the Gap columns. Since the nuclear norm measures the overall magnitude of gradient updates; thus, a smaller value suggests the model requires less energy to adapt to high-quality data. It further indicates that high-quality data should be aligned with the learned knowledge of pretrained LLMs. 
At the same time, high-quality subsets also yield consistently larger effective ranks in their gradients represented by the large positive values (in green) for the Gap columns. 
A higher value suggests that more update directions are activated, which means high-quality data leads to richer, more multi-dimensional parameter updates, which likely improves the model’s ability to generalize and capture nuanced features of the instruction pairs.

\textbf{Our finding illustrates that multiple different definitions of ``data quality'' can converge to overarching gradient properties, revealing a unified view by gradient-based spectral analysis.}


\subsubsection{Effects of Original Dataset Qualities}

\begin{figure*}[t]
    \centering
    \includegraphics[width=\linewidth]{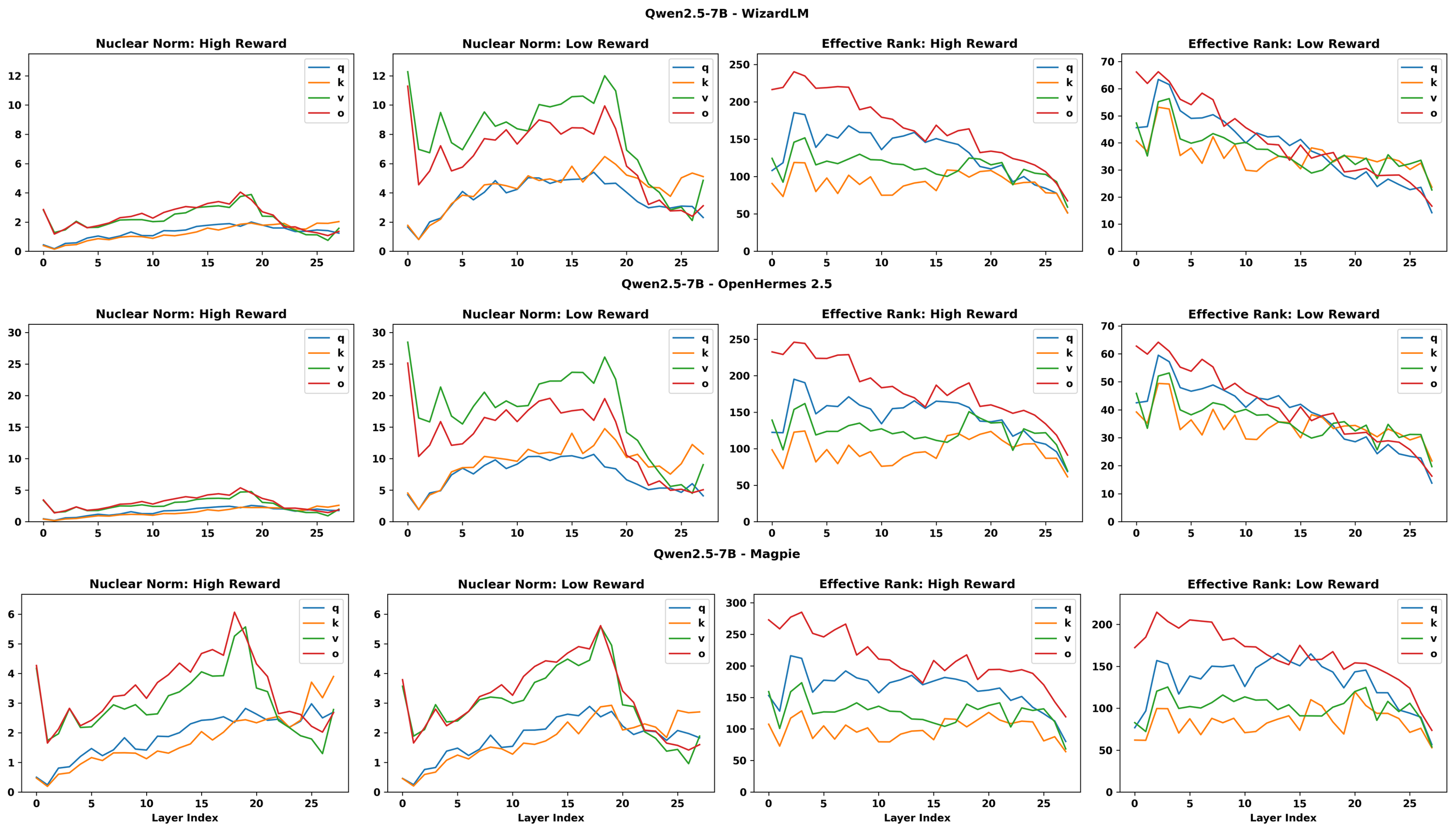}
    \vspace{-2mm}
    \caption{Low/high-quality data (measured by Reward) and their gradient properties (nuclear norms and effective ranks) across layers on diverse datasets including WizardLM, OpenHermes 2.5, and Magpie. 
    The y-axis scales are kept the same for nuclear norms, while different for effective ranks, due to the large discrepancy. 
    \textbf{For each specific model, the shapes of the gradient curves derived from different data sources are almost the same. 
    The nuclear norm fails to reflect the quality discrepancies between datasets, while the effective ranks still works promisingly, e.g., Magpie has higher rank than others. }
    }
    \label{fig:412}
    \vspace{-2mm}
\end{figure*}


In this section, we aim to investigate the potential ineffectiveness of gradient properties for distinguishing high- or low-quality data. Similar to the previous settings, we split the data from different sources, WizardLM~\citep{xu2023wizardlm}, OpenHermes 2.5~\citep{OpenHermes2.5}, and Magpie ~\citep{xu2024magpiealignmentdatasynthesis}, into high- and low-quality subsets, calculating their effects on gradients based on Qwen2.5-7B. The nuclear norm and effective rank changes with layer indexes are shown in Figure \ref{fig:412}. The shapes of gradient curves are almost kept the same for one specific model, whatever the data quality. 

Moreover, we find that when an entire dataset is already composed of fairly clean and coherent instruction–response samples, e.g., the Magpie dataset shown in the third row, nuclear norms offer limited discriminative power in distinguishing it further, represented by the similar scales for both types of data. 
We hypothesize that once instructions and responses cross a certain threshold of clarity and consistency (i.e., minimal noise or confusion), the gradient magnitudes needed to adapt to these subsets no longer diverge sharply. 
In other words, the nuclear norm may effectively ``saturate'' and stop registering small but meaningful differences among these relatively high-quality groups.
On the contrary, effective ranks remain sensitive to smaller quality disparities, even within a dataset already recognized for solid instruction–response fidelity.
Consequently, effective rank acts as a finer-grained lens, revealing that while both subsets are indeed ``good'' enough to produce stable gradients, the higher-quality examples still manage to activate a broader range of update directions.

\subsection{Reasoning Data}

\subsubsection{Unifying Quality Evaluation of Instruction and Reasoning Data}

\begin{figure*}[t]
    \centering
    \includegraphics[width=\linewidth]{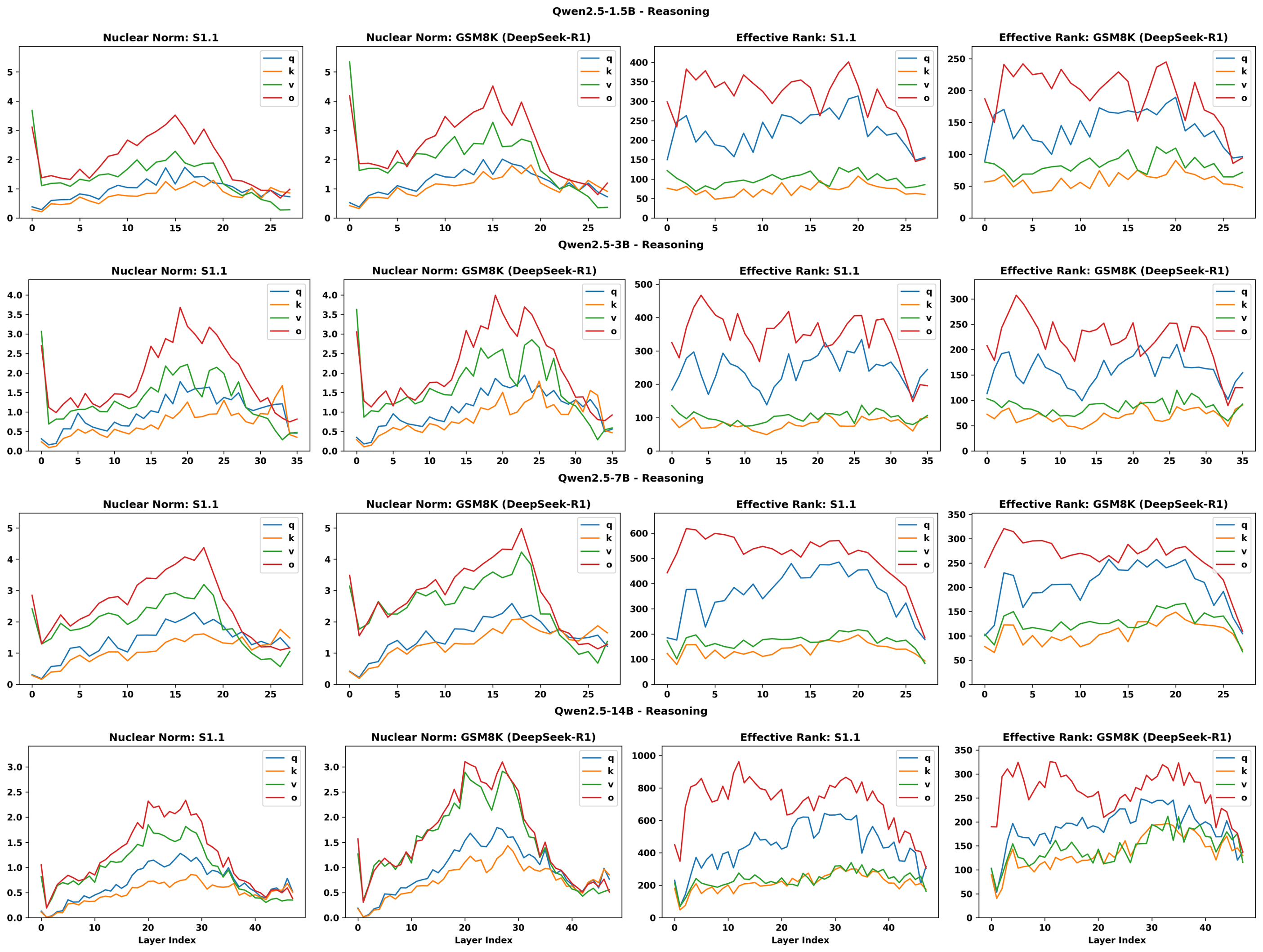}
    \vspace{-6mm}
    \caption{Model size scaling law for gradient properties. 
    \textbf{Within the same model family, the layer-wise gradient statistics and dynamics are relatively consistent. Gradients on larger models exhibit better capabilities to distinguish data quality}, revealed by the increasing y-axis scales from the $1.5$B model to $14$B model. }
    \label{fig:422}
    \vspace{-2mm}
\end{figure*}
\begin{table*}[t]
\centering
\resizebox{\textwidth}{!}{%
\begin{tabular}{l|l|
                c c c c|
                c c c c|
                c c c c|
                c c c c}
\toprule
\multirow{2}{*}{\textbf{Dataset}} & \multirow{2}{*}{\textbf{Metrics}} 
& \multicolumn{8}{c|}{\textbf{Nuclear Norm}}
& \multicolumn{8}{c}{\textbf{Effective Rank}} \\
\cmidrule(lr){3-6}\cmidrule(lr){7-10}\cmidrule(lr){11-14}\cmidrule(lr){15-18}
 & & \textbf{Proj} & High & Low & Gap 
   & \textbf{Proj} & High & Low & Gap
   & \textbf{Proj} & High & Low & Gap
   & \textbf{Proj} & High & Low & Gap \\
\midrule
\multirow{2}{*}{\textbf{Magpie}} & \multirow{2}{*}{\textbf{Difficulty}}
  & k & 1.7 & 1.7 & \cellcolor{red!20}{0.0}
  & q & 1.8 & 1.8 & \cellcolor{red!20}{0.0}
  & k & 95.9 & 83.8 & \cellcolor{green!20}{12.1}
  & q & 153.3 & 124.4 & \cellcolor{green!20}{28.9} \\
& 
  & v & 3.0 & 3.1 & \cellcolor{red!20}{-0.1}
  & o & 3.3 & 3.3 & \cellcolor{red!20}{0.0}
  & v & 118.0 & 102.7 & \cellcolor{green!20}{15.3}
  & o & 195.1 & 151.7 & \cellcolor{green!20}{43.4} \\
\midrule



\multirow{2}{*}{\textbf{Reasoning}} & \multirow{2}{*}{\textbf{Difficulty}}
  & k & 1.0 & 1.3 & \cellcolor{red!20}{-0.3}
  & q & 1.3 & 1.5 & \cellcolor{red!20}{-0.2}
  & k & 138.8 & 106.1 & \cellcolor{green!20}{32.7}
  & q & 361.2 & 203.3 & \cellcolor{green!20}{157.9} \\
& 
  & v & 1.9 & 2.4 & \cellcolor{red!20}{-0.5}
  & o & 2.5 & 2.8 & \cellcolor{red!20}{-0.3}
  & v & 170.4 & 126.7 & \cellcolor{green!20}{43.7}
  & o & 509.9 & 263.1 & \cellcolor{green!20}{246.8} \\

\bottomrule
\end{tabular}
}
\vspace{-2mm}
\caption{Comparing the gradient properties between instruction data vs. reasoning data. For the reasoning data, \textit{High} denotes data sampled from s1.1, and \textit{Low} denotes sampled from GSM8K with DeepSeek-R1 responses. Reasoning data shows lower nuclear norms and higher effective ranks compared with instruction data. 
\textbf{Our analysis of gradients unifies the quality evaluation for reasoning and instruction data on both higher effective ranks reflecting higher quality. Moreover, the metric distinguishes low-/high- quality data by large gaps.} }
\vspace{-4mm}
\label{tab:421}
\end{table*}










Beyond analyzing general instruction-following data, the recent surge in reasoning models encourages us to further explore the effects of reasoning data on LLM gradients. 
Motivated by recent success on eliciting models' reasoning capabilities by distilling from stronger reasoning models through simple SFT, e.g., s1~\citep{muennighoff2025s1simpletesttimescaling}, LIMO~\citep{ye2025limo}, DeepSeek-R1 Distilled Qwen~\citep{deepseekai2025deepseekr1incentivizingreasoningcapability}, we formulate the exploration on reasoning data in the same structure as on instruction-following data. 
Compared with general instruction-following data, a key difference of advanced reasoning data is the utilization of dynamic long CoTs and the accordance with the test-time scaling law. 
To distinguish this data from general instruction-following data, we notate them as reasoning data. 
As illustrated by s1, $1,000$ difficult math problems are sufficient to elicit LLMs' reasoning capabilities; thus, we ask: \textbf{Does reasoning data have similar effects to gradients with previous general instruction data? Can the quality of advanced reasoning data be further distinguished by gradient properties like nuclear norm and effective rank?}

To formulate our experiments, we utilize the s1.1 data as the high-quality subset and GSM8K~\citep{cobbe2021trainingverifierssolvemath} (responses also generated by DeepSeek-R1~\citep{ deepseekai2025deepseekr1incentivizingreasoningcapability}) as the low-quality subset for our reasoning data experiments. 
The results are shown in Table \ref{tab:421}, in which the gradients are calculated based on Qwen2.5-7B, comparing the general instruction-following data with the reasoning data. Surprisingly, our experiments demonstrate that the same gradient-derived metrics, nuclear norms and effective ranks, remain applicable for the advanced reasoning data: 
(i) \textit{Reasoning vs. Instruction data}: Reasoning data, even the lower-quality subset, leads to lower nuclear norms and higher effective ranks compared with previous high-quality instruction data, suggesting the much higher data quality of recent reasoning data. 
(ii) \textit{Higher- vs. Lower-quality reasoning data}: Even for the reasoning data with supreme data quality, there still exists a consistent and unified trend with the previous instruction data, i.e., smaller nuclear norms and larger effective ranks for higher-quality data. 
Moreover, the gaps in effective ranks between high- and low-quality subsets are more pronounced for reasoning data than for general instruction data, which might provide a potential explanation on why s1.1 can reach such a promising performance with only $1000$ data. 

\textbf{We are the first to investigate and compare the effects of general instruction-following and reasoning data toward LLM gradients in the training process. 
We reveal that both data types can be \textit{unified} into a consistent pattern of gradient-based signals regarding quality, providing a unified view for understanding the quality effects. }

\subsubsection{Effects of Model Sizes}

\begin{figure*}[t]
    \centering
    \includegraphics[width=\linewidth]{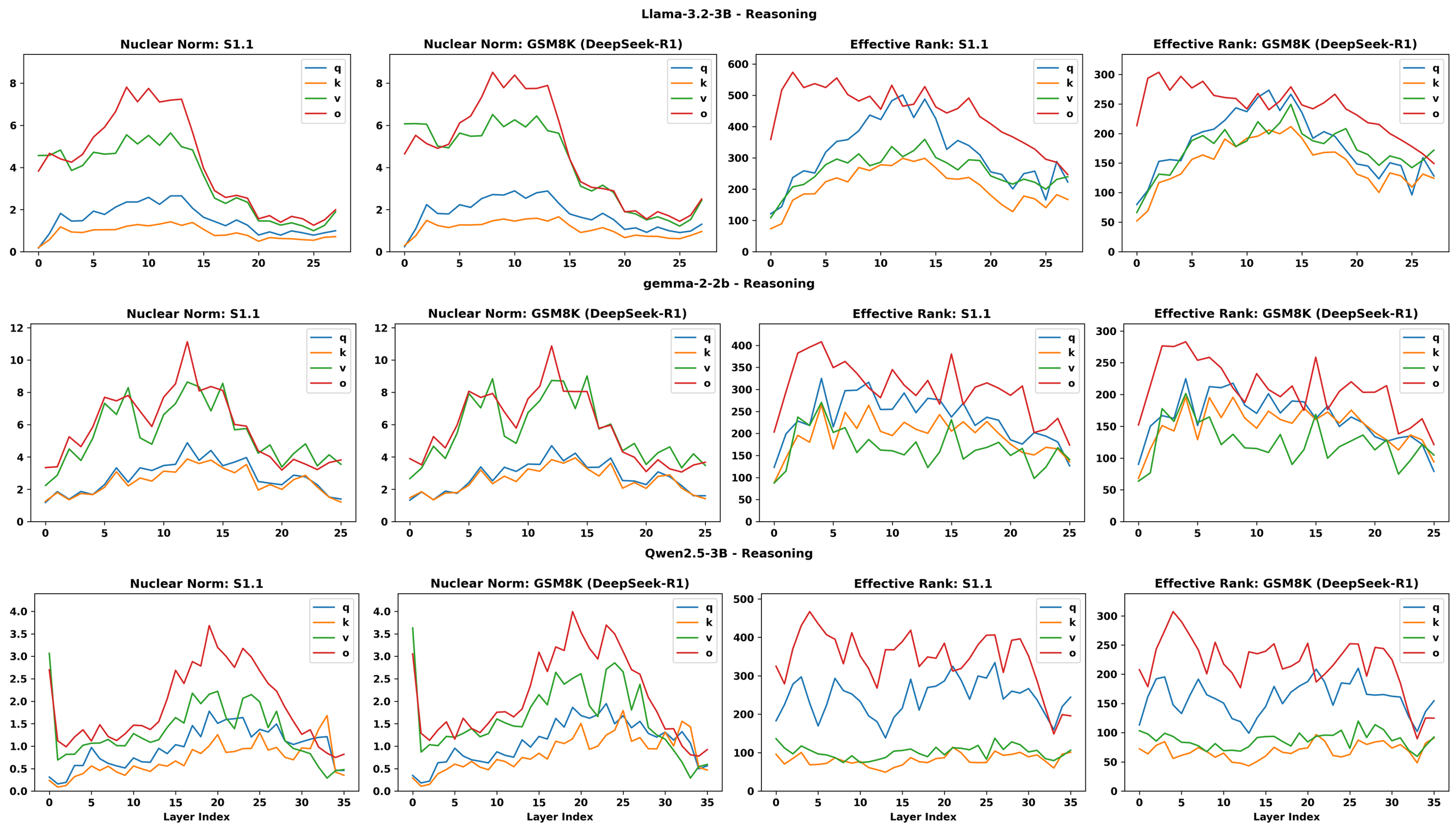}
    \vspace{-2mm}
    \caption{Gradient properties across different model families. \textbf{The gradient dynamics of the same data on different model families are largely different. This might be caused by their distinct model structures or training recipes and may reflect their different capabilities.} }
    \label{fig:423}
    \vspace{-4mm}
\end{figure*}


Figure \ref{fig:422} presents the gradient curves of LLMs with different sizes in the Qwen2.5 family on our reasoning data. 
The shapes of the gradient curves for reasoning data are almost the same as the instruction-following data, which further verifies our unified view on both types of data. 
Moreover, a consistent observation is that the overall shape of these layer-wise curves remains relatively (not strictly) stable as we move from smaller to larger models. 
For instance, GSM8K-based fine-tuning triggers a peak in nuclear norm at the mid layers for Qwen2.5-1.5B, a qualitatively similar, albeit scaled, peak in Qwen2.5-14B also can be identified. 

Moreover, another interesting finding is that larger models tend to amplify the distinction between high- and low-quality subsets. Specifically, the gaps in effective rank between s1.1 and GSM8K grow accordingly when moving to bigger models. In other words, the larger model is more sensitive to whether the provided reasoning path is coherent and informative. 

\subsubsection{Effects of Model Families}
\begin{table*}[t]
\centering
\resizebox{\textwidth}{!}{%
\begin{tabular}{l|l|
                c c c|
                c c c|
                c c c|
                c c c}
\toprule
\multirow{2}{*}{\textbf{Dataset}} & \multirow{2}{*}{\textbf{Metric}} 
& \multicolumn{6}{c|}{\textbf{Same-layer Similarity}}
& \multicolumn{6}{c}{\textbf{Adjacent-layer Similarity}} \\
\cmidrule(lr){3-5}\cmidrule(lr){6-8}\cmidrule(lr){9-11}\cmidrule(lr){12-14}
 & & \textbf{Proj} & High & Low  
   & \textbf{Proj} & High & Low
   & \textbf{Proj} & High & Low
   & \textbf{Proj} & High & Low\\
\midrule
\multirow{2}{*}{\textbf{Magpie}} & \multirow{2}{*}{\textbf{Difficulty}}
  & \multirow{2}{*}{k - v} & \multirow{2}{*}{-0.7e-3} & \multirow{2}{*}{-0.7e-3}
  & \multirow{2}{*}{q - o} & \multirow{2}{*}{0.0e-3} & \multirow{2}{*}{0.0e-3}
  & k & -0.1e-3 & -0.1e-3
  & q & 0.1e-3 & -0.1e-3 \\
& 
  &  &  &  
  &  &  &  
  & v & 0.3e-3 & 0.1e-3 
  & o & 0.4e-3 & 0.3e-3  \\
\midrule

\multirow{2}{*}{\textbf{Reasoning}} & \multirow{2}{*}{\textbf{Difficulty}}
  & \multirow{2}{*}{k - v} & \multirow{2}{*}{0.0e-3} & \multirow{2}{*}{0.0e-3} 
  & \multirow{2}{*}{q - o} & \multirow{2}{*}{0.1e-3} & \multirow{2}{*}{-0.9e-3} 
  & k & -2.0e-3 & -2.0e-3 
  & q & 0.0e-3 & 0.2e-3  \\
& 
  &  &  & 
  &  &  & 
  & v & 1.0e-3 & 2.0e-3
  & o & 1.0e-3 & 2.0e-3\\

\bottomrule
\end{tabular}
}
\vspace{-2mm}
\caption{Gradient similarity metrics remain excessively small and cannot reflect the differences between instruction/reasoning data of low/high-quality. \textbf{It shows that layer-wise gradients in LLM post-training are nearly orthogonal, indicating that the similarity of gradients is not an effective indicator of data quality.}  }
\vspace{-4mm}
\label{tab:520}
\end{table*}


In this section, we broaden the scope to compare entirely different model families, including Qwen2.5-3B, Llama3.2-3B, and gemma2-2B, each possessing distinct pretraining recipes and architectural configurations. 
The first notable observation is that the layer-by-layer ``shape'' of the gradients can vary significantly among families. 
For instance, Llama3.2-3B might exhibit consistently higher nuclear norms in its early layers compared to Qwen2.5-3B, reflecting differences in embedding or attention initialization. 
Despite these baseline discrepancies, the relative gap between high- and low-quality reasoning data persists across all families. 
In other words, regardless of how each model is architected or pre-trained, high-quality data still yields smaller nuclear norms and larger effective ranks.
This cross-family analysis suggests that reasoning data can be a valuable resource regardless of the specific LLM architectures. 
At the same time, our analysis shows the existence of family-specific ``fingerprints'' for nuclear norms and effective ranks, reflecting architectural and pretraining differences, which might potentially be useful for a better understanding of the LLM architectures. 

\subsubsection{Similarity-based Metrics}

While our SVD-based metrics are proved consistently effective at capturing data-quality differences, similarity-based metrics, namely, same-layer similarity and adjacent-layer similarity, do not appear to yield meaningful signals in our experiments, as shown in Table \ref{tab:520}.
We keep the values in the same magnitude for easier comparison. 
In the table, we compare similarity measures for both instruction and reasoning data across high- and low-quality subsets. Regardless of dataset type or quality level, the reported cosine similarities remain extremely close to zero, with minimal observable variation.
These low similarities suggest that the gradients for LLM SFT are nearly orthogonal, indicating that similarity on gradients is not an effective indicator of data quality.

\section{Conclusion}

We introduce a unified gradient-based framework for analyzing how varying data quality, ranging from general instruction-following to reasoning data, shapes the finetuning of LLMs. 
By examining the layer-wise gradients, we show that different quality metrics converge on remarkably similar gradient signatures, specifically, smaller gradient magnitudes and broader gradient directions for high-quality data. 
Notably, this pattern holds across multiple model families and parameter scales 
and, for the first time, reveals how reasoning data induce even higher effective ranks and thus richer parameter updates. 

\clearpage
\section*{Limitation}

The main limitation of this work is not conduct further experiments on whether these gradient spectral statistics can be directly utilized for follow-up data selection. 
However, as mentioned in the paper, the main contribution of this work lies in the first investigation on the unified effects of data quality metrics on gradients, rather than utilizing it as a new data selection method. 

\clearpage
\bibliography{colm2025_conference}

\appendix
\clearpage

\startcontents[appendix]
\printcontents[appendix]{ }{0}{\section*{Table of Contents}}

\section{Related Work}

\subsection{Gradient Analysis}

Empirical analyses of gradient dynamics have provided deep insights into how large language models (LLMs) learn during both pretraining and fine-tuning. A recurring observation is that LLM training tends to concentrate updates in a low-dimensional subspace, where only a small fraction of parameters account for the majority of the gradient magnitude~\citep{song2024sparsefinetuningpretrainedlarge}. This sparsity in meaningful directions underlies the effectiveness of methods like parameter-efficient fine-tuning and low-rank adaptation. For example, \citet{li2025enhancinglargelanguagemodel} propose gradient masking to zero out low-impact updates and concentrate learning on parameters with high gradient magnitude. These observations are consistent with findings from spectral studies of the Hessian~\citep{sagun2018empiricalanalysishessianoverparametrized, ghorbani2019investigationneuralnetoptimization}, which reveal that deep networks—including transformers—tend to learn in a space governed by a few dominant gradient directions. This low effective dimensionality has been linked to both generalization and the robustness of optimization trajectories~\citep{li2018measuringintrinsicdimensionobjective}.

Gradient-based perspectives have also proven valuable in understanding how different types of training data influence learning. In curriculum learning, for instance, it has been observed that ordering data from easy to hard helps align gradients in earlier phases of training, resulting in smoother convergence and better generalization~\citep{hacohen2019powercurriculumlearningtraining}. In multi-task and continual learning, techniques like GradNorm~\citep{chen2018gradnormgradientnormalizationadaptive} and PCGrad~\citep{yu2020gradientsurgerymultitasklearning} explicitly manipulate gradient magnitudes and directions to resolve conflicts between tasks. Furthermore, gradient-based influence functions~\citep{koh2020understandingblackboxpredictionsinfluence, pruthi2020estimatingtrainingdatainfluence} have been used to trace how individual data points affect model updates and predictions, enabling researchers to detect noisy or highly influential examples. In the context of instruction tuning and alignment, such analyses can reveal how instruction-following or reasoning-rich data differentially steer the model's behavior.

A closely related work by~\citet{li2024happenedllmslayerstrained} studies how cognitive styles (fast vs. slow thinking) impact LLM gradient dynamics, showing that reasoning-rich (slow thinking) data yields smaller, stable gradients compared to direct-answer (fast thinking) data. Extending this, we introduce new metrics: effective rank for gradient complexity, and two similarity metrics (Same-layer Similarityl, Adjacent-layer Similarity) to analyze gradient alignment within and between layers. These metrics provide deeper insights into how instruction-following and reasoning-rich data influence LLM training.


\subsection{SFT and Data Selection}

Supervised fine-tuning (SFT) is the subsequent step where an LLM is further trained on labeled examples to specialize it for specific tasks or to align it with human instructions. This process updates the model’s weights using task-specific input–output pairs (e.g., prompts and desired responses)~\citep{wang2023aligning}
InstructGPT~\citep{ouyang2022training} demonstrated that fine-tuning GPT-3 on human-curated instruction datasets yields models that more effectively follow user instructions compared to the base models. This approach, referred to as instruction tuning, enhances performance and exhibits robust generalizability.

In the instruction tuning, the concept ``quality is all you need'' is widely accepted~\citep{zhou2023lima, touvron2023llama2, havrilla2024surveyingeffectsqualitydiversity}.
Earlier research focused on curating high-quality datasets through human experts or powerful LLMs~\citep{khashabi-etal-2020-unifiedqa, ye-etal-2021-crossfit, wei2022finetuned, wang-etal-2022-super, du-etal-2022-glm, alpaca}. Although these approaches yield high-quality datasets, they are time-consuming, expensive, and offer limited diversity. 

However, LIMA~\citep{zhou2023lima} demonstrates that as few as 1,000 high-quality, human-curated training instances can substantially enhance instruction-following performance. Building on this work, data selection has emerged as an increasingly critical stage in instruction tuning.
InsTag~\citep{lu2023instag} uses a powerful proprietary model to tag samples within SFT datasets based on semantics and intentions, subsequently selecting data that meet predefined criteria for complexity and diversity in their tags. 
Alpagasus~\citep{chen2023alpagasus} utilizes proprietary LLMs chatGPT and Claude2 to automatically identify and filter out low-quality data within the dataset. 
Cherry LLM~\citep{cherry} introduces Instruction-Following Difficulty (IFD) scores—a self-guided metric for evaluating instruction difficulty without relying on additional LLMs—and uses these scores to select datasets.
Motivated by Humpback~\citep{li2023self}, Selective Reflection-Tuning \citep{Li2024SelectiveRS} introduces a teacher-student collaborative pipeline that is guided by the IFD score and its reverse version to select the data based on the evaluation feasibility.
\citet{du2023mods} and \citet{bukharin2023data} utilize reward models as the metric for measuring data quality and subsequently select the data. 
DEITA~\citep{liu2023makes} employs ChatGPT to diversify datasets, then applies various data selection metrics to construct a high-quality dataset.
Superfiltering~\citep{Li2024SuperfilteringWD} demonstrates that both weak and strong language models consistently assess instruction difficulty, thereby streamlining the filtering process. 
Instruct Mining~\citep{caoinstruction}  presents a method for automatically selecting high-quality instruction-following data using natural language quality indicators.
SelectIT~\citep{liu2024selectit} proposes an uncertainty-aware self-filtering approach that leverages an LLM’s intrinsic uncertainty to select high-quality instruction-tuning data
LESS~\citep{xia2024less} constructs a low-dimensional ``gradient datastore'' for candidate data, subsequently selecting examples whose gradient influence closely matches that of a limited set of target demonstration examples.
Collectively, these studies focus on distinguishing high-quality data samples from lower-quality ones for effective instruction tuning.

\subsection{Reasoning Capability}

Recent advancements in large language model architectures and training objectives have led to a paradigm shift from fast, heuristic (System-1) response generation to deliberate, multi-step (System-2) reasoning processes \citep{ahn-etal-2024-large, li2024happenedllmslayerstrained, jin2024impactreasoningsteplength, besta2025reasoninglanguagemodelsblueprint, li202512surveyreasoning, chen2025reasoningerasurveylong, li-etal-2025-understanding, li2025schoenfeld, xiong2025deliberate, xiongenhancing}, marked by advanced reasoning models such as OpenAI's o1/o3 \citep{openai2024openaio1card}, QwQ \citep{qwen2025qwen25technicalreport}, and DeepSeek-R1 \citep{deepseekai2025deepseekr1incentivizingreasoningcapability}. System 1 models (e.g., GPT-4o \citep{openai2024gpt4ocard}, LLaMA-3 \citep{dubey2024llama3herdmodels}, DeepSeek-V3 \citep{deepseekai2025deepseekv3technicalreport}) generate intuitive, rapid responses but often struggle with more intricate tasks. In contrast, System 2 models adopt methodical analysis and iterative self-critique, which strengthens their performance on complex reasoning benchmarks.


Various post-training methods have been introduced to further enhance these capabilities. A number of studies employ reinforcement learning strategies to guide models toward superior reasoning approaches~\citep{shao2024deepseekmathpushinglimitsmathematical, cui2025processreinforcementimplicitrewards}. Moreover, researchers have demonstrated that instruction tuning on meticulously curated, high-quality datasets can greatly boost a model’s reasoning performance~\citep{ye2025limo, muennighoff2025s1simpletesttimescaling}. However, despite the advancement in reasoning capability, the issue of overthinking \citep{chen2025think23overthinkingo1like, fan2025missingpremiseexacerbatesoverthinking, qu2025surveyefficientreasoninglarge, liu2025efficientinferencelargereasoning} brings further challenges to the area. 

\section{Prompts for Data Evaluation}

For our selected data quality evaluation metrics, InsTag and Difficulty are the prompt-based methods. We use GPT4o for the evaluation and the prompts are provided in Figure \ref{appendix_prompt_instag} and Figure \ref{appendix_prompt_difficulty}.

\begin{figure*}[h]
  \centering
  \parbox{0.98\textwidth}{
        \rule{0.98\textwidth}{1.5pt} 
        Prompt for InsTag \\
        \rule{0.98\textwidth}{0.8pt} 
        \textbf{System Prompt} \\
        You are a helpful assistant. \\

        \textbf{User Prompt} \\
        You are a tagging system that provides useful tags for instruction intentions to distinguish instructions for a helpful AI assistant. Below is an instruction:\\
        \text{[begin]}\\
        \{\textit{instruction}\}\\
        \text{[end]}\\
        Please provide coarse-grained tags, such as "Spelling and Grammar Check" and "Cosplay", to identify the main intentions of the above instruction. Your answer should be a list that includes the titles of tags and a brief explanation of each tag. You can provide several tags as you wish.
        Your response has to strictly follow this JSON format: \text{[\{"tag": str, "explanation": str\},\{"tag": str, "explanation": str\},...]}. Please respond in English.\\
        \rule{0.98\textwidth}{0.8pt} 
  }
\caption{
The prompt for InsTag.
} 
\label{appendix_prompt_instag} 
\end{figure*}

\begin{figure*}[h]
  \centering
  \parbox{0.98\textwidth}{
        \rule{0.98\textwidth}{1.5pt} 
        Prompt for Difficulty \\
        \rule{0.98\textwidth}{0.8pt} 
        \textbf{System Prompt} \\
        You are a helpful assistant. \\

        \textbf{User Prompt} \\
        You are a difficulty estimation system that can rate the difficulty level of instruction intentions. Below is an instruction:\\
        \text{[begin]}\\
        \{\textit{instruction}\}\\
        \text{[end]}\\
        The instruction can be tagged with a difficulty level from 1 to 10, where 1 is the easiest and 10 is the hardest. Please rate the difficulty level of the instruction. 
        Please first output a single line containing the difficulty score. Then, provide a brief explanation of why you rated the instruction with that difficulty score.\\
        \rule{0.98\textwidth}{0.8pt} 
  }
\caption{
The prompt for Difficulty.
} 
\label{appendix_prompt_difficulty} 
\end{figure*}

\clearpage
\section{Gradient Curves for Different Models and Metrics}

\subsection{Qwen2.5 1.5B}

\begin{table*}[t]
\centering
\resizebox{\textwidth}{!}{%
\begin{tabular}{l|l|
                c c c c|
                c c c c|
                c c c c|
                c c c c}
\toprule
\multirow{2}{*}{\textbf{Dataset}} & \multirow{2}{*}{\textbf{Metrics}} 
& \multicolumn{8}{c|}{\textbf{Nuclear Norm}}
& \multicolumn{8}{c}{\textbf{Effective Rank}} \\
\cmidrule(lr){3-6}\cmidrule(lr){7-10}\cmidrule(lr){11-14}\cmidrule(lr){15-18}
 & & \textbf{Proj} & High & Low & Gap 
   & \textbf{Proj} & High & Low & Gap
   & \textbf{Proj} & High & Low & Gap
   & \textbf{Proj} & High & Low & Gap \\
\midrule
\multirow{8}{*}{\textbf{WizardLM}}& \multirow{2}{*}{\textbf{Difficulty}}  & k & 1.6 & 4.2 & \cellcolor{red!20}{-2.6}  & q & 1.6 & 3.4 & \cellcolor{red!20}{-1.7}  & k & 56.2 & 14.0 & \cellcolor{green!20}{42.2}  & q & 104.5 & 16.4 & \cellcolor{green!20}{88.1} \\
&   & v & 2.7 & 7.9 & \cellcolor{red!20}{-5.2}  & o & 3.2 & 6.5 & \cellcolor{red!20}{-3.3}  & v & 81.2 & 16.9 & \cellcolor{green!20}{64.4}  & o & 142.3 & 18.3 & \cellcolor{green!20}{124.0} \\\cmidrule(lr){2-18}
 & \multirow{2}{*}{\textbf{IFD (GPT2)}}  & k & 1.8 & 4.7 & \cellcolor{red!20}{-3.0}  & q & 1.8 & 3.6 & \cellcolor{red!20}{-1.8}  & k & 44.4 & 15.7 & \cellcolor{green!20}{28.7}  & q & 74.7 & 17.3 & \cellcolor{green!20}{57.5} \\
&   & v & 3.2 & 7.9 & \cellcolor{red!20}{-4.7}  & o & 3.6 & 6.7 & \cellcolor{red!20}{-3.2}  & v & 62.9 & 18.6 & \cellcolor{green!20}{44.3}  & o & 96.6 & 19.9 & \cellcolor{green!20}{76.8} \\\cmidrule(lr){2-18}
 & \multirow{2}{*}{\textbf{InsTag}}  & k & 1.6 & 5.0 & \cellcolor{red!20}{-3.4}  & q & 1.7 & 4.0 & \cellcolor{red!20}{-2.4}  & k & 57.9 & 15.0 & \cellcolor{green!20}{42.8}  & q & 108.9 & 17.8 & \cellcolor{green!20}{91.1} \\
&   & v & 2.7 & 9.1 & \cellcolor{red!20}{-6.4}  & o & 3.2 & 7.6 & \cellcolor{red!20}{-4.3}  & v & 83.5 & 18.5 & \cellcolor{green!20}{65.0}  & o & 148.8 & 20.1 & \cellcolor{green!20}{128.7} \\\cmidrule(lr){2-18}
 & \multirow{2}{*}{\textbf{Reward}}  & k & 1.1 & 4.4 & \cellcolor{red!20}{-3.3}  & q & 1.2 & 3.8 & \cellcolor{red!20}{-2.6}  & k & 56.8 & 24.7 & \cellcolor{green!20}{32.1}  & q & 103.9 & 30.1 & \cellcolor{green!20}{73.8} \\
&   & v & 1.9 & 7.8 & \cellcolor{red!20}{-5.8}  & o & 2.3 & 7.1 & \cellcolor{red!20}{-4.7}  & v & 80.7 & 30.3 & \cellcolor{green!20}{50.4}  & o & 140.4 & 35.2 & \cellcolor{green!20}{105.2} \\\bottomrule
\end{tabular}
}
\caption{Qwen2.5-1.5B - WizardLM - SVD-based Metrics}
\label{tab:qwen2.5-1.5b_wizardlm_nuclear_erank}
\end{table*}

\begin{table*}[t]
\centering
\resizebox{\textwidth}{!}{%
\begin{tabular}{l|l|
                c c c|
                c c c|
                c c c|
                c c c}
\toprule
\multirow{2}{*}{\textbf{Dataset}} & \multirow{2}{*}{\textbf{Metric}} 
& \multicolumn{6}{c|}{\textbf{Same-layer Similarity}}
& \multicolumn{6}{c}{\textbf{Adjacent-layer Similarity}} \\
\cmidrule(lr){3-5}\cmidrule(lr){6-8}\cmidrule(lr){9-11}\cmidrule(lr){12-14}
 & & \textbf{Proj} & High & Low  
   & \textbf{Proj} & High & Low
   & \textbf{Proj} & High & Low
   & \textbf{Proj} & High & Low\\
\midrule
\multirow{8}{*}{\textbf{WizardLM}} & \multirow{2}{*}{\textbf{Difficulty}}  & \multirow{2}{*}{k - v} & \multirow{2}{*}{-6.9e-04} & \multirow{2}{*}{-1.3e-03}  & \multirow{2}{*}{q - o} & \multirow{2}{*}{1.8e-06} & \multirow{2}{*}{7.1e-06}  & k & 6.0e-04 & 6.1e-04  & q & -1.6e-04 & 4.9e-04 \\
&   &  &  &   &  &  &   & v & 3.9e-04 & -8.1e-04  & o & -7.3e-04 & 2.2e-03 \\\cmidrule(lr){2-14}
  & \multirow{2}{*}{\textbf{IFD (GPT2)}}  & \multirow{2}{*}{k - v} & \multirow{2}{*}{-7.0e-04} & \multirow{2}{*}{-9.9e-04}  & \multirow{2}{*}{q - o} & \multirow{2}{*}{-4.3e-06} & \multirow{2}{*}{3.6e-06}  & k & 2.5e-04 & 3.6e-04  & q & 1.3e-04 & 5.9e-04 \\
&   &  &  &   &  &  &   & v & 3.8e-04 & -9.5e-05  & o & -5.5e-04 & 3.5e-03 \\\cmidrule(lr){2-14}
  & \multirow{2}{*}{\textbf{InsTag}}  & \multirow{2}{*}{k - v} & \multirow{2}{*}{-1.2e-03} & \multirow{2}{*}{-6.4e-04}  & \multirow{2}{*}{q - o} & \multirow{2}{*}{1.1e-05} & \multirow{2}{*}{1.1e-05}  & k & 1.1e-03 & 2.4e-04  & q & 2.3e-04 & 6.9e-04 \\
&   &  &  &   &  &  &   & v & 2.9e-04 & -3.0e-05  & o & -8.6e-04 & 3.7e-03 \\\cmidrule(lr){2-14}
  & \multirow{2}{*}{\textbf{Reward}}  & \multirow{2}{*}{k - v} & \multirow{2}{*}{-4.7e-04} & \multirow{2}{*}{-6.6e-04}  & \multirow{2}{*}{q - o} & \multirow{2}{*}{-9.3e-06} & \multirow{2}{*}{-7.0e-06}  & k & 7.5e-04 & 1.1e-03  & q & 7.5e-05 & -7.6e-05 \\
&   &  &  &   &  &  &   & v & 3.7e-04 & -1.9e-04  & o & -1.4e-04 & 1.6e-03 \\\bottomrule
\end{tabular}
}
\caption{Qwen2.5-1.5B - WizardLM - Similarity-based Metrics}
\label{tab:qwen2.5-1.5b_wizardlm_cosine}
\end{table*}

\begin{table*}[t]
\centering
\resizebox{\textwidth}{!}{%
\begin{tabular}{l|l|
                c c c c|
                c c c c|
                c c c c|
                c c c c}
\toprule
\multirow{2}{*}{\textbf{Dataset}} & \multirow{2}{*}{\textbf{Metrics}} 
& \multicolumn{8}{c|}{\textbf{Nuclear Norm}}
& \multicolumn{8}{c}{\textbf{Effective Rank}} \\
\cmidrule(lr){3-6}\cmidrule(lr){7-10}\cmidrule(lr){11-14}\cmidrule(lr){15-18}
 & & \textbf{Proj} & High & Low & Gap 
   & \textbf{Proj} & High & Low & Gap
   & \textbf{Proj} & High & Low & Gap
   & \textbf{Proj} & High & Low & Gap \\
\midrule
\multirow{8}{*}{\textbf{OpenHermes 2.5}}& \multirow{2}{*}{\textbf{Difficulty}}  & k & 2.0 & 4.2 & \cellcolor{red!20}{-2.2}  & q & 2.0 & 3.4 & \cellcolor{red!20}{-1.5}  & k & 55.2 & 21.1 & \cellcolor{green!20}{34.1}  & q & 101.2 & 26.6 & \cellcolor{green!20}{74.6} \\
&   & v & 3.4 & 7.3 & \cellcolor{red!20}{-4.0}  & o & 3.8 & 6.6 & \cellcolor{red!20}{-2.8}  & v & 79.4 & 26.6 & \cellcolor{green!20}{52.9}  & o & 136.5 & 30.5 & \cellcolor{green!20}{106.0} \\\cmidrule(lr){2-18}
 & \multirow{2}{*}{\textbf{IFD (GPT2)}}  & k & 1.6 & 11.0 & \cellcolor{red!20}{-9.4}  & q & 1.7 & 8.7 & \cellcolor{red!20}{-7.0}  & k & 49.0 & 21.0 & \cellcolor{green!20}{28.0}  & q & 85.7 & 24.3 & \cellcolor{green!20}{61.4} \\
&   & v & 3.0 & 18.8 & \cellcolor{red!20}{-15.8}  & o & 3.5 & 16.3 & \cellcolor{red!20}{-12.8}  & v & 69.3 & 25.1 & \cellcolor{green!20}{44.1}  & o & 110.7 & 28.9 & \cellcolor{green!20}{81.8} \\\cmidrule(lr){2-18}
 & \multirow{2}{*}{\textbf{InsTag}}  & k & 1.6 & 4.1 & \cellcolor{red!20}{-2.5}  & q & 1.7 & 3.4 & \cellcolor{red!20}{-1.7}  & k & 57.6 & 20.3 & \cellcolor{green!20}{37.4}  & q & 107.6 & 25.7 & \cellcolor{green!20}{81.9} \\
&   & v & 2.8 & 7.1 & \cellcolor{red!20}{-4.4}  & o & 3.3 & 6.5 & \cellcolor{red!20}{-3.2}  & v & 82.5 & 25.8 & \cellcolor{green!20}{56.8}  & o & 145.2 & 29.9 & \cellcolor{green!20}{115.3} \\\cmidrule(lr){2-18}
 & \multirow{2}{*}{\textbf{Reward}}  & k & 1.2 & 8.7 & \cellcolor{red!20}{-7.5}  & q & 1.4 & 6.9 & \cellcolor{red!20}{-5.6}  & k & 58.6 & 24.0 & \cellcolor{green!20}{34.6}  & q & 108.5 & 30.6 & \cellcolor{green!20}{77.9} \\
&   & v & 2.2 & 15.4 & \cellcolor{red!20}{-13.2}  & o & 2.8 & 13.4 & \cellcolor{red!20}{-10.6}  & v & 84.2 & 29.5 & \cellcolor{green!20}{54.8}  & o & 147.1 & 36.2 & \cellcolor{green!20}{110.9} \\\bottomrule
\end{tabular}
}
\caption{Qwen2.5-1.5B - OpenHermes 2.5 - SVD-based Metrics}
\label{tab:qwen2.5-1.5b_openhermes_2.5_nuclear_erank}
\end{table*}

\begin{table*}[t]
\centering
\resizebox{\textwidth}{!}{%
\begin{tabular}{l|l|
                c c c|
                c c c|
                c c c|
                c c c}
\toprule
\multirow{2}{*}{\textbf{Dataset}} & \multirow{2}{*}{\textbf{Metric}} 
& \multicolumn{6}{c|}{\textbf{Same-layer Similarity}}
& \multicolumn{6}{c}{\textbf{Adjacent-layer Similarity}} \\
\cmidrule(lr){3-5}\cmidrule(lr){6-8}\cmidrule(lr){9-11}\cmidrule(lr){12-14}
 & & \textbf{Proj} & High & Low  
   & \textbf{Proj} & High & Low
   & \textbf{Proj} & High & Low
   & \textbf{Proj} & High & Low\\
\midrule
\multirow{8}{*}{\textbf{OpenHermes 2.5}} & \multirow{2}{*}{\textbf{Difficulty}}  & \multirow{2}{*}{k - v} & \multirow{2}{*}{-4.1e-04} & \multirow{2}{*}{-5.2e-04}  & \multirow{2}{*}{q - o} & \multirow{2}{*}{8.9e-06} & \multirow{2}{*}{1.1e-05}  & k & 6.8e-04 & -3.4e-04  & q & 7.3e-05 & 5.5e-05 \\
&   &  &  &   &  &  &   & v & 4.4e-04 & 1.1e-04  & o & -4.8e-04 & 1.7e-03 \\\cmidrule(lr){2-14}
  & \multirow{2}{*}{\textbf{IFD (GPT2)}}  & \multirow{2}{*}{k - v} & \multirow{2}{*}{-1.1e-03} & \multirow{2}{*}{6.3e-04}  & \multirow{2}{*}{q - o} & \multirow{2}{*}{-1.0e-05} & \multirow{2}{*}{5.2e-06}  & k & 1.5e-04 & 6.4e-04  & q & -1.3e-04 & 7.9e-05 \\
&   &  &  &   &  &  &   & v & 6.5e-05 & -4.0e-05  & o & -5.4e-04 & 2.3e-03 \\\cmidrule(lr){2-14}
  & \multirow{2}{*}{\textbf{InsTag}}  & \multirow{2}{*}{k - v} & \multirow{2}{*}{-7.8e-04} & \multirow{2}{*}{-1.5e-04}  & \multirow{2}{*}{q - o} & \multirow{2}{*}{1.2e-06} & \multirow{2}{*}{2.2e-06}  & k & 1.7e-03 & -2.0e-04  & q & 1.4e-04 & 4.0e-04 \\
&   &  &  &   &  &  &   & v & -3.1e-04 & 2.0e-04  & o & -6.1e-04 & 2.0e-03 \\\cmidrule(lr){2-14}
  & \multirow{2}{*}{\textbf{Reward}}  & \multirow{2}{*}{k - v} & \multirow{2}{*}{-3.0e-04} & \multirow{2}{*}{-5.8e-04}  & \multirow{2}{*}{q - o} & \multirow{2}{*}{-4.1e-07} & \multirow{2}{*}{-3.8e-06}  & k & 6.1e-04 & 1.1e-03  & q & -3.0e-04 & -2.1e-04 \\
&   &  &  &   &  &  &   & v & 2.2e-06 & -1.8e-04  & o & -1.2e-03 & 3.0e-03 \\\bottomrule
\end{tabular}
}
\caption{Qwen2.5-1.5B - OpenHermes 2.5 - Similarity-based Metrics}
\label{tab:qwen2.5-1.5b_openhermes_2.5_cosine}
\end{table*}

\begin{table*}[t]
\centering
\resizebox{\textwidth}{!}{%
\begin{tabular}{l|l|
                c c c c|
                c c c c|
                c c c c|
                c c c c}
\toprule
\multirow{2}{*}{\textbf{Dataset}} & \multirow{2}{*}{\textbf{Metrics}} 
& \multicolumn{8}{c|}{\textbf{Nuclear Norm}}
& \multicolumn{8}{c}{\textbf{Effective Rank}} \\
\cmidrule(lr){3-6}\cmidrule(lr){7-10}\cmidrule(lr){11-14}\cmidrule(lr){15-18}
 & & \textbf{Proj} & High & Low & Gap 
   & \textbf{Proj} & High & Low & Gap
   & \textbf{Proj} & High & Low & Gap
   & \textbf{Proj} & High & Low & Gap \\
\midrule
\multirow{8}{*}{\textbf{Magpie}}& \multirow{2}{*}{\textbf{Difficulty}}  & k & 1.5 & 1.5 & \cellcolor{green!20}{-0.0}  & q & 1.6 & 1.6 & \cellcolor{green!20}{-0.0}  & k & 55.8 & 51.7 & \cellcolor{green!20}{4.1}  & q & 110.0 & 92.3 & \cellcolor{green!20}{17.7} \\
&   & v & 2.6 & 2.8 & \cellcolor{red!20}{-0.1}  & o & 3.2 & 3.3 & \cellcolor{red!20}{-0.1}  & v & 78.6 & 73.2 & \cellcolor{green!20}{5.4}  & o & 149.2 & 121.8 & \cellcolor{green!20}{27.4} \\\cmidrule(lr){2-18}
 & \multirow{2}{*}{\textbf{IFD (GPT2)}}  & k & 1.4 & 1.2 & \cellcolor{green!20}{0.2}  & q & 1.6 & 1.2 & \cellcolor{green!20}{0.3}  & k & 58.5 & 55.7 & \cellcolor{green!20}{2.8}  & q & 117.5 & 99.1 & \cellcolor{green!20}{18.4} \\
&   & v & 2.6 & 2.1 & \cellcolor{green!20}{0.5}  & o & 3.3 & 2.5 & \cellcolor{green!20}{0.8}  & v & 83.6 & 77.3 & \cellcolor{green!20}{6.3}  & o & 158.7 & 135.4 & \cellcolor{green!20}{23.3} \\\cmidrule(lr){2-18}
 & \multirow{2}{*}{\textbf{InsTag}}  & k & 1.5 & 1.5 & \cellcolor{green!20}{0.0}  & q & 1.6 & 1.6 & \cellcolor{green!20}{0.1}  & k & 61.6 & 52.1 & \cellcolor{green!20}{9.5}  & q & 124.1 & 96.7 & \cellcolor{green!20}{27.5} \\
&   & v & 2.6 & 2.7 & \cellcolor{green!20}{-0.0}  & o & 3.2 & 3.2 & \cellcolor{green!20}{0.0}  & v & 86.8 & 73.9 & \cellcolor{green!20}{13.0}  & o & 170.5 & 128.9 & \cellcolor{green!20}{41.6} \\\cmidrule(lr){2-18}
 & \multirow{2}{*}{\textbf{Reward}}  & k & 1.4 & 1.4 & \cellcolor{green!20}{-0.0}  & q & 1.6 & 1.5 & \cellcolor{green!20}{0.1}  & k & 58.1 & 48.8 & \cellcolor{green!20}{9.3}  & q & 118.7 & 94.9 & \cellcolor{green!20}{23.8} \\
&   & v & 2.5 & 2.5 & \cellcolor{red!20}{-0.1}  & o & 3.2 & 2.9 & \cellcolor{green!20}{0.3}  & v & 84.4 & 67.5 & \cellcolor{green!20}{16.9}  & o & 164.2 & 128.3 & \cellcolor{green!20}{35.9} \\\bottomrule
\end{tabular}
}
\caption{Qwen2.5-1.5B - Magpie - SVD-based Metrics}
\label{tab:qwen2.5-1.5b_magpie_nuclear_erank}
\end{table*}

\begin{table*}[t]
\centering
\resizebox{\textwidth}{!}{%
\begin{tabular}{l|l|
                c c c|
                c c c|
                c c c|
                c c c}
\toprule
\multirow{2}{*}{\textbf{Dataset}} & \multirow{2}{*}{\textbf{Metric}} 
& \multicolumn{6}{c|}{\textbf{Same-layer Similarity}}
& \multicolumn{6}{c}{\textbf{Adjacent-layer Similarity}} \\
\cmidrule(lr){3-5}\cmidrule(lr){6-8}\cmidrule(lr){9-11}\cmidrule(lr){12-14}
 & & \textbf{Proj} & High & Low  
   & \textbf{Proj} & High & Low
   & \textbf{Proj} & High & Low
   & \textbf{Proj} & High & Low\\
\midrule
\multirow{8}{*}{\textbf{Magpie}} & \multirow{2}{*}{\textbf{Difficulty}}  & \multirow{2}{*}{k - v} & \multirow{2}{*}{-2.4e-03} & \multirow{2}{*}{-1.5e-03}  & \multirow{2}{*}{q - o} & \multirow{2}{*}{-1.1e-05} & \multirow{2}{*}{7.9e-06}  & k & -2.7e-04 & 3.6e-04  & q & -2.0e-04 & -6.8e-05 \\
&   &  &  &   &  &  &   & v & 1.5e-03 & 3.3e-04  & o & 1.4e-05 & -1.1e-03 \\\cmidrule(lr){2-14}
  & \multirow{2}{*}{\textbf{IFD (GPT2)}}  & \multirow{2}{*}{k - v} & \multirow{2}{*}{-1.6e-03} & \multirow{2}{*}{-1.6e-03}  & \multirow{2}{*}{q - o} & \multirow{2}{*}{1.2e-05} & \multirow{2}{*}{-1.2e-05}  & k & -6.7e-04 & 1.4e-03  & q & -1.4e-04 & -6.0e-05 \\
&   &  &  &   &  &  &   & v & 3.1e-04 & 2.2e-04  & o & -9.4e-04 & -1.1e-04 \\\cmidrule(lr){2-14}
  & \multirow{2}{*}{\textbf{InsTag}}  & \multirow{2}{*}{k - v} & \multirow{2}{*}{-1.3e-03} & \multirow{2}{*}{-1.5e-03}  & \multirow{2}{*}{q - o} & \multirow{2}{*}{-4.0e-06} & \multirow{2}{*}{1.0e-05}  & k & 1.5e-03 & -2.2e-04  & q & -6.5e-05 & 3.0e-04 \\
&   &  &  &   &  &  &   & v & -4.2e-05 & 1.1e-03  & o & -2.7e-04 & -1.5e-03 \\\cmidrule(lr){2-14}
  & \multirow{2}{*}{\textbf{Reward}}  & \multirow{2}{*}{k - v} & \multirow{2}{*}{-1.5e-03} & \multirow{2}{*}{-3.3e-03}  & \multirow{2}{*}{q - o} & \multirow{2}{*}{-1.5e-05} & \multirow{2}{*}{7.8e-06}  & k & 6.1e-04 & -6.4e-04  & q & -3.0e-04 & -6.0e-04 \\
&   &  &  &   &  &  &   & v & 6.0e-04 & 2.0e-03  & o & -2.1e-03 & -1.2e-03 \\\bottomrule
\end{tabular}
}
\caption{Qwen2.5-1.5B - Magpie - Similarity-based Metrics}
\label{tab:qwen2.5-1.5b_magpie_cosine}
\end{table*}

\begin{table*}[t]
\centering
\resizebox{\textwidth}{!}{%
\begin{tabular}{l|l|
                c c c c|
                c c c c|
                c c c c|
                c c c c}
\toprule
\multirow{2}{*}{\textbf{Dataset}} & \multirow{2}{*}{\textbf{Metrics}} 
& \multicolumn{8}{c|}{\textbf{Nuclear Norm}}
& \multicolumn{8}{c}{\textbf{Effective Rank}} \\
\cmidrule(lr){3-6}\cmidrule(lr){7-10}\cmidrule(lr){11-14}\cmidrule(lr){15-18}
 & & \textbf{Proj} & High & Low & Gap 
   & \textbf{Proj} & High & Low & Gap
   & \textbf{Proj} & High & Low & Gap
   & \textbf{Proj} & High & Low & Gap \\
\midrule
\multirow{2}{*}{\textbf{Reasoning}}& \multirow{2}{*}{\textbf{Reasoning}}  & k & 0.8 & 1.1 & \cellcolor{red!20}{-0.3}  & q & 1.0 & 1.2 & \cellcolor{red!20}{-0.3}  & k & 72.1 & 59.5 & \cellcolor{green!20}{12.6}  & q & 223.8 & 141.7 & \cellcolor{green!20}{82.1} \\
&   & v & 1.4 & 2.0 & \cellcolor{red!20}{-0.6}  & o & 2.0 & 2.5 & \cellcolor{red!20}{-0.5}  & v & 98.7 & 82.6 & \cellcolor{green!20}{16.1}  & o & 311.7 & 194.2 & \cellcolor{green!20}{117.5} \\
\bottomrule
\end{tabular}
}
\caption{Qwen2.5-1.5B - Reasoning - SVD-based Metrics}
\label{tab:qwen2.5-1.5b_reasoning_nuclear_erank}
\end{table*}

\begin{table*}[t]
\centering
\resizebox{\textwidth}{!}{%
\begin{tabular}{l|l|
                c c c|
                c c c|
                c c c|
                c c c}
\toprule
\multirow{2}{*}{\textbf{Dataset}} & \multirow{2}{*}{\textbf{Metric}} 
& \multicolumn{6}{c|}{\textbf{Same-layer Similarity}}
& \multicolumn{6}{c}{\textbf{Adjacent-layer Similarity}} \\
\cmidrule(lr){3-5}\cmidrule(lr){6-8}\cmidrule(lr){9-11}\cmidrule(lr){12-14}
 & & \textbf{Proj} & High & Low  
   & \textbf{Proj} & High & Low
   & \textbf{Proj} & High & Low
   & \textbf{Proj} & High & Low\\
\midrule
\multirow{2}{*}{\textbf{Reasoning}} & \multirow{2}{*}{\textbf{Reasoning}}  & \multirow{2}{*}{k - v} & \multirow{2}{*}{4.8e-04} & \multirow{2}{*}{6.6e-04}  & \multirow{2}{*}{q - o} & \multirow{2}{*}{-4.2e-06} & \multirow{2}{*}{-1.2e-05}  & k & 1.6e-03 & 3.6e-04  & q & 5.3e-04 & -1.2e-04 \\
&   &  &  &   &  &  &   & v & 9.1e-04 & 4.2e-03  & o & 5.2e-04 & -9.7e-04 \\
\bottomrule
\end{tabular}
}
\caption{Qwen2.5-1.5B - Reasoning - Similarity-based Metrics}
\label{tab:qwen2.5-1.5b_reasoning_cosine}
\end{table*}

\begin{figure*}[t]
    \centering
    \includegraphics[width=1\textwidth]{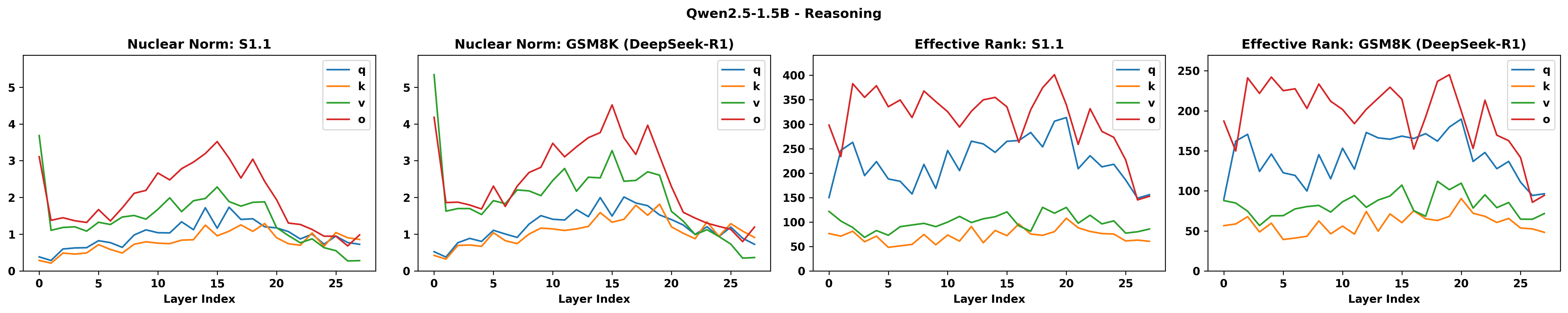}
    \caption{Qwen2.5 1.5B - Reasoning Data}
    \label{fig:qwen25_15b_reasoning}
\end{figure*}

\begin{figure*}[t]
    \centering
    \includegraphics[width=1\textwidth]{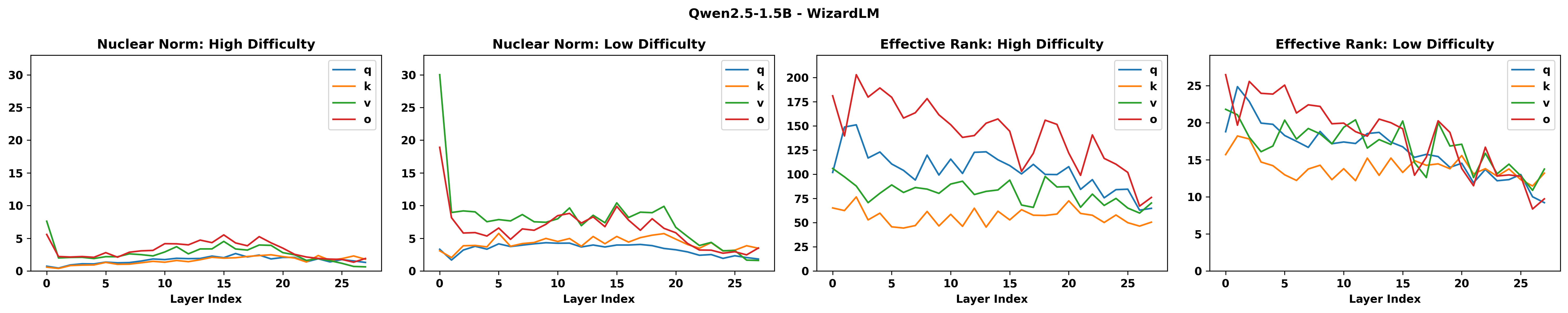}
    \caption{Qwen2.5 1.5B - WizardLM with Difficulty Metric}
    \label{fig:qwen25_15b_wiz_diff}
\end{figure*}

\begin{figure*}[t]
    \centering
    \includegraphics[width=1\textwidth]{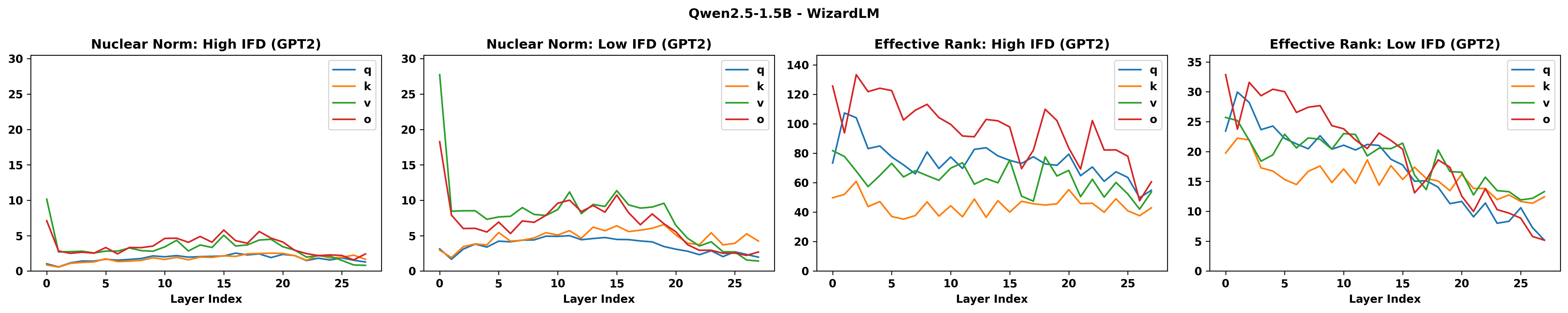}
    \caption{Qwen2.5 1.5B - WizardLM with IFD (GPT-2) Metric}
    \label{fig:qwen25_15b_wiz_ifd}
\end{figure*}

\begin{figure*}[t]
    \centering
    \includegraphics[width=1\textwidth]{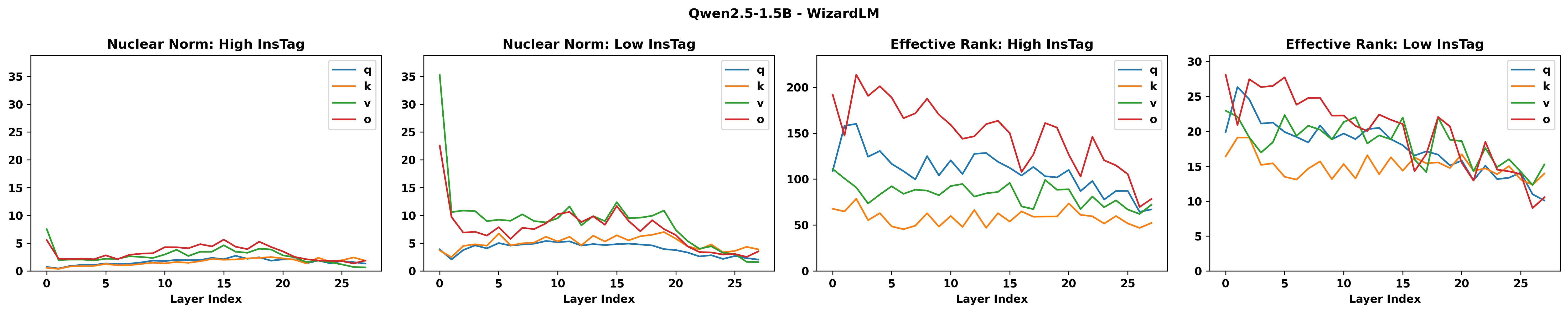}
    \caption{Qwen2.5 1.5B - WizardLM with InsTag Metric}
    \label{fig:qwen25_15b_wiz_instag}
\end{figure*}

\begin{figure*}[t]
    \centering
    \includegraphics[width=1\textwidth]{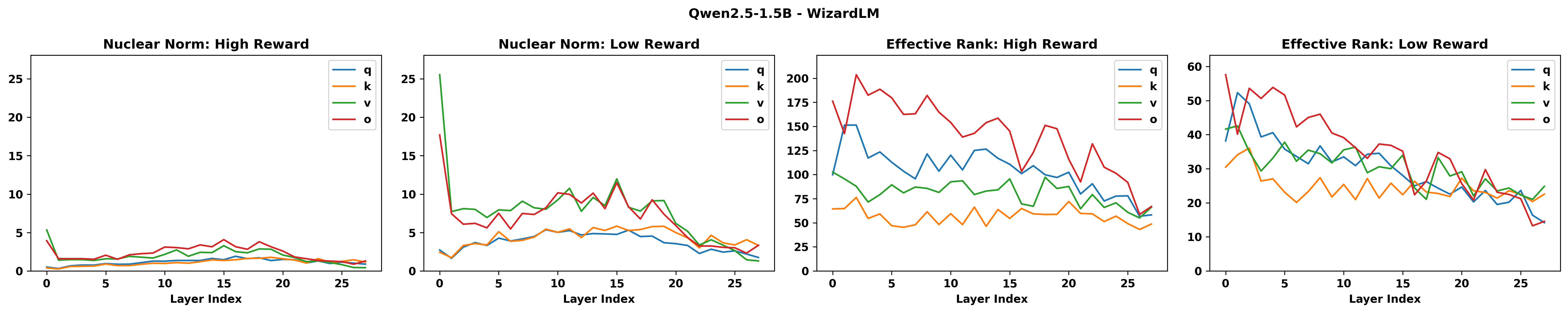}
    \caption{Qwen2.5 1.5B - WizardLM with Reward Model Metric}
    \label{fig:qwen25_15b_wiz_reward}
\end{figure*}

\begin{figure*}[t]
    \centering
    \includegraphics[width=1\textwidth]{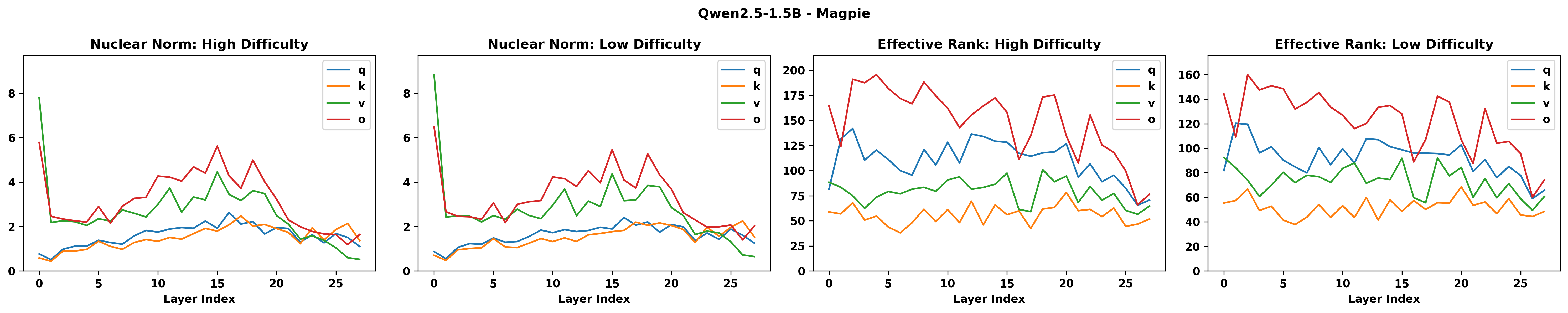}
    \caption{Qwen2.5 1.5B - Magpie with Difficulty Metric}
    \label{fig:qwen25_15b_mag_diff}
\end{figure*}

\begin{figure*}[t]
    \centering
    \includegraphics[width=1\textwidth]{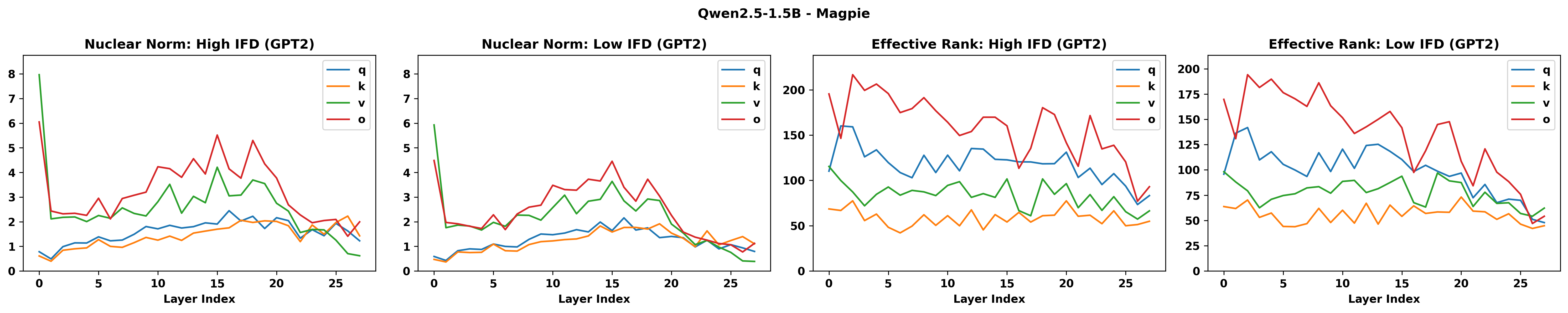}
    \caption{Qwen2.5 1.5B - Magpie with IFD (GPT-2) Metric}
    \label{fig:qwen25_15b_mag_ifd}
\end{figure*}

\begin{figure*}[t]
    \centering
    \includegraphics[width=1\textwidth]{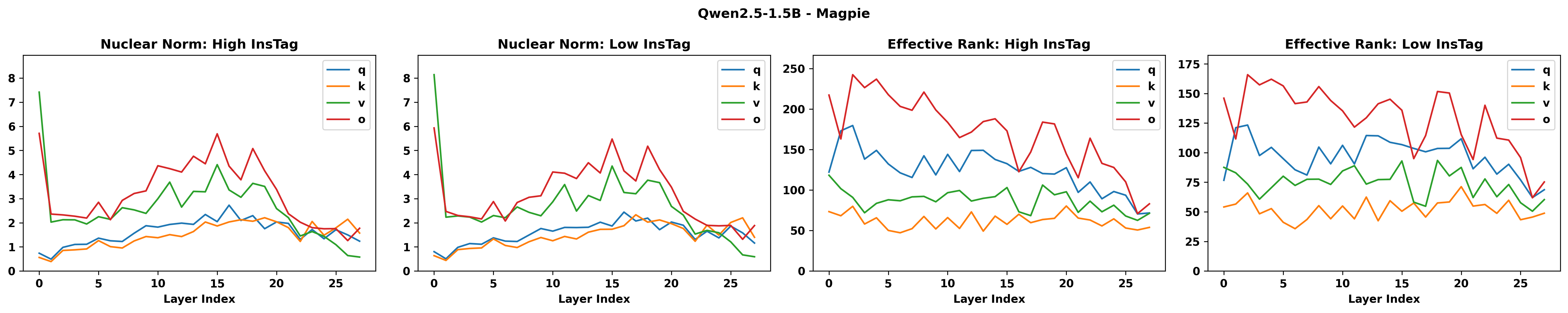}
    \caption{Qwen2.5 1.5B - Magpie with InsTag Metric}
    \label{fig:qwen25_15b_mag_instag}
\end{figure*}

\begin{figure*}[t]
    \centering
    \includegraphics[width=1\textwidth]{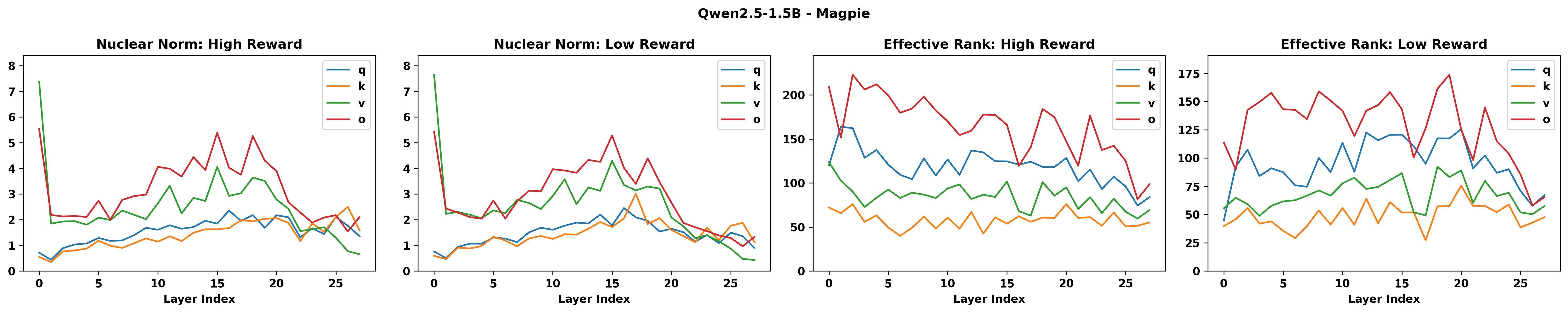}
    \caption{Qwen2.5 1.5B - Magpie with Reward Model Metric}
    \label{fig:qwen25_15b_mag_reward}
\end{figure*}

\begin{figure*}[t]
    \centering
    \includegraphics[width=1\textwidth]{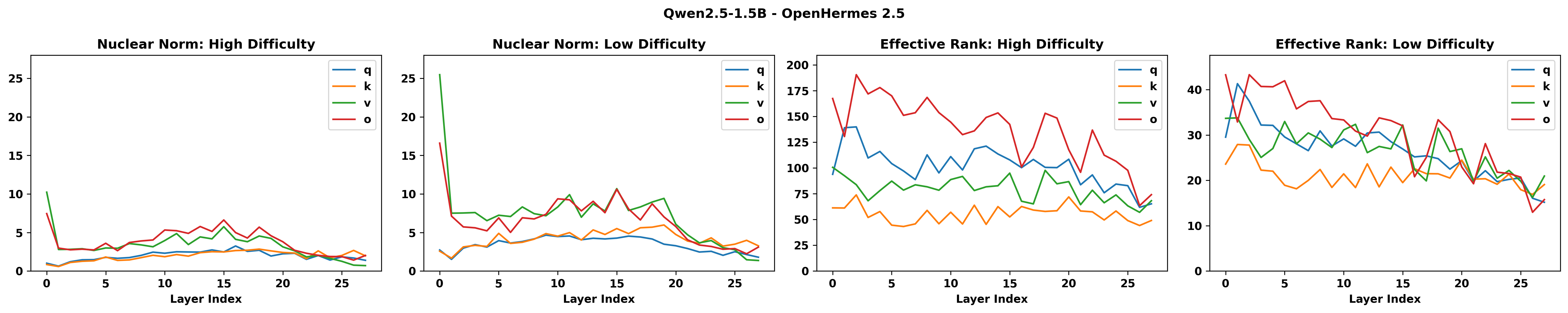}
    \caption{Qwen2.5 1.5B - OpenHermes with Difficulty Metric}
    \label{fig:qwen25_15b_her_diff}
\end{figure*}

\begin{figure*}[t]
    \centering
    \includegraphics[width=1\textwidth]{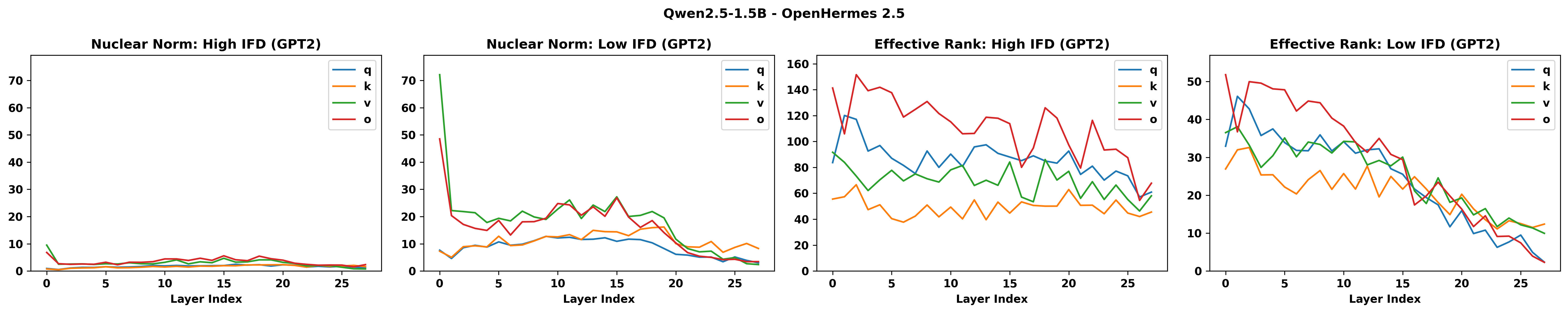}
    \caption{Qwen2.5 1.5B - OpenHermes with IFD (GPT-2) Metric}
    \label{fig:qwen25_15b_her_ifd}
\end{figure*}

\begin{figure*}[t]
    \centering
    \includegraphics[width=1\textwidth]{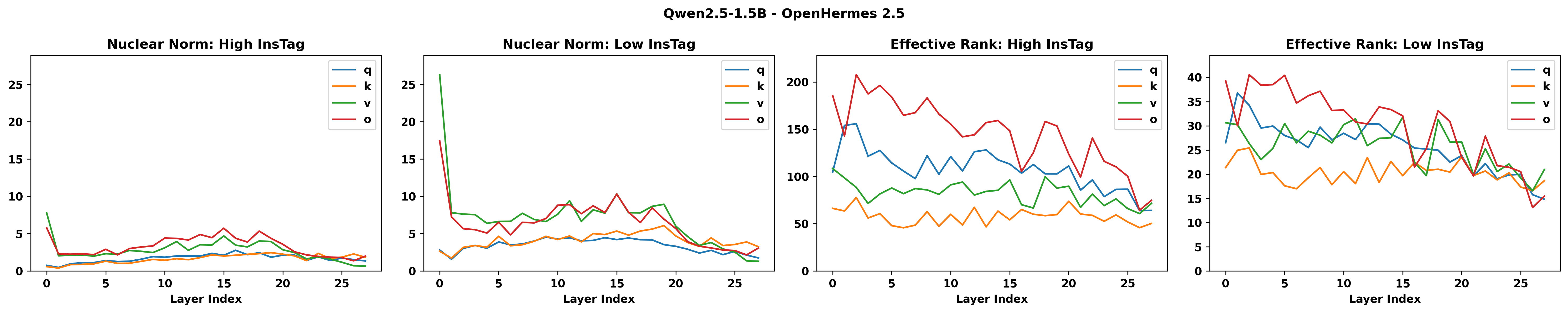}
    \caption{Qwen2.5 1.5B - OpenHermes with InsTag Metric}
    \label{fig:qwen25_15b_her_instag}
\end{figure*}

\begin{figure*}[t]
    \centering
    \includegraphics[width=1\textwidth]{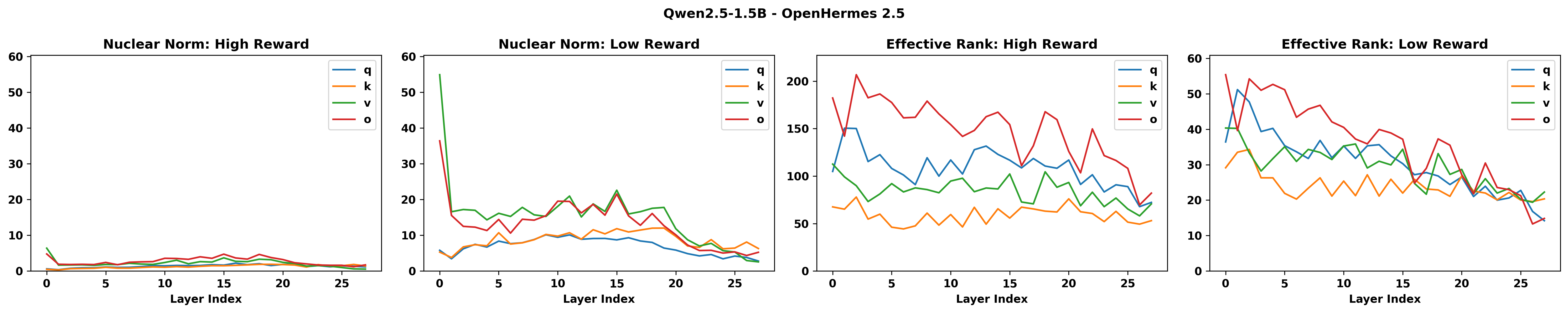}
    \caption{Qwen2.5 1.5B - OpenHermes with Reward Model Metric}
    \label{fig:qwen25_15b_her_reward}
\end{figure*}

\clearpage
\subsection{Qwen2.5 3B}

\begin{table*}[t]
\centering
\resizebox{\textwidth}{!}{%
\begin{tabular}{l|l|
                c c c c|
                c c c c|
                c c c c|
                c c c c}
\toprule
\multirow{2}{*}{\textbf{Dataset}} & \multirow{2}{*}{\textbf{Metrics}} 
& \multicolumn{8}{c|}{\textbf{Nuclear Norm}}
& \multicolumn{8}{c}{\textbf{Effective Rank}} \\
\cmidrule(lr){3-6}\cmidrule(lr){7-10}\cmidrule(lr){11-14}\cmidrule(lr){15-18}
 & & \textbf{Proj} & High & Low & Gap 
   & \textbf{Proj} & High & Low & Gap
   & \textbf{Proj} & High & Low & Gap
   & \textbf{Proj} & High & Low & Gap \\
\midrule
\multirow{8}{*}{\textbf{WizardLM}}& \multirow{2}{*}{\textbf{Difficulty}}  & k & 1.2 & 2.9 & \cellcolor{red!20}{-1.7}  & q & 1.3 & 2.5 & \cellcolor{red!20}{-1.2}  & k & 65.6 & 15.9 & \cellcolor{green!20}{49.7}  & q & 118.7 & 18.0 & \cellcolor{green!20}{100.7} \\
&   & v & 2.1 & 5.7 & \cellcolor{red!20}{-3.6}  & o & 2.5 & 4.9 & \cellcolor{red!20}{-2.4}  & v & 87.3 & 18.3 & \cellcolor{green!20}{69.1}  & o & 160.4 & 19.8 & \cellcolor{green!20}{140.7} \\\cmidrule(lr){2-18}
 & \multirow{2}{*}{\textbf{IFD (GPT2)}}  & k & 1.3 & 3.2 & \cellcolor{red!20}{-2.0}  & q & 1.4 & 2.8 & \cellcolor{red!20}{-1.4}  & k & 51.9 & 18.2 & \cellcolor{green!20}{33.8}  & q & 84.0 & 19.5 & \cellcolor{green!20}{64.5} \\
&   & v & 2.4 & 5.7 & \cellcolor{red!20}{-3.3}  & o & 2.7 & 5.0 & \cellcolor{red!20}{-2.3}  & v & 68.6 & 20.3 & \cellcolor{green!20}{48.3}  & o & 106.3 & 22.0 & \cellcolor{green!20}{84.3} \\\cmidrule(lr){2-18}
 & \multirow{2}{*}{\textbf{InsTag}}  & k & 1.2 & 3.6 & \cellcolor{red!20}{-2.4}  & q & 1.3 & 3.2 & \cellcolor{red!20}{-1.9}  & k & 67.2 & 17.1 & \cellcolor{green!20}{50.2}  & q & 123.8 & 19.5 & \cellcolor{green!20}{104.3} \\
&   & v & 2.1 & 6.8 & \cellcolor{red!20}{-4.8}  & o & 2.5 & 6.0 & \cellcolor{red!20}{-3.5}  & v & 89.5 & 19.7 & \cellcolor{green!20}{69.8}  & o & 168.2 & 21.5 & \cellcolor{green!20}{146.7} \\\cmidrule(lr){2-18}
 & \multirow{2}{*}{\textbf{Reward}}  & k & 0.8 & 3.2 & \cellcolor{red!20}{-2.4}  & q & 0.9 & 3.0 & \cellcolor{red!20}{-2.1}  & k & 66.0 & 28.8 & \cellcolor{green!20}{37.1}  & q & 119.3 & 34.0 & \cellcolor{green!20}{85.3} \\
&   & v & 1.5 & 6.0 & \cellcolor{red!20}{-4.6}  & o & 1.8 & 5.5 & \cellcolor{red!20}{-3.8}  & v & 86.9 & 33.2 & \cellcolor{green!20}{53.6}  & o & 161.7 & 38.6 & \cellcolor{green!20}{123.1} \\\bottomrule
\end{tabular}
}
\caption{Qwen2.5-3B - WizardLM - SVD-based Metrics}
\label{tab:qwen2.5-3b_wizardlm_nuclear_erank}
\end{table*}

\begin{table*}[t]
\centering
\resizebox{\textwidth}{!}{%
\begin{tabular}{l|l|
                c c c|
                c c c|
                c c c|
                c c c}
\toprule
\multirow{2}{*}{\textbf{Dataset}} & \multirow{2}{*}{\textbf{Metric}} 
& \multicolumn{6}{c|}{\textbf{Same-layer Similarity}}
& \multicolumn{6}{c}{\textbf{Adjacent-layer Similarity}} \\
\cmidrule(lr){3-5}\cmidrule(lr){6-8}\cmidrule(lr){9-11}\cmidrule(lr){12-14}
 & & \textbf{Proj} & High & Low  
   & \textbf{Proj} & High & Low
   & \textbf{Proj} & High & Low
   & \textbf{Proj} & High & Low\\
\midrule
\multirow{8}{*}{\textbf{WizardLM}} & \multirow{2}{*}{\textbf{Difficulty}}  & \multirow{2}{*}{k - v} & \multirow{2}{*}{-5.6e-04} & \multirow{2}{*}{-5.4e-04}  & \multirow{2}{*}{q - o} & \multirow{2}{*}{9.1e-06} & \multirow{2}{*}{-3.9e-06}  & k & -1.3e-03 & -7.8e-04  & q & 2.5e-05 & 1.0e-04 \\
&   &  &  &   &  &  &   & v & -5.1e-04 & -6.4e-04  & o & 1.4e-03 & 2.1e-03 \\\cmidrule(lr){2-14}
  & \multirow{2}{*}{\textbf{IFD (GPT2)}}  & \multirow{2}{*}{k - v} & \multirow{2}{*}{-2.7e-04} & \multirow{2}{*}{-8.4e-04}  & \multirow{2}{*}{q - o} & \multirow{2}{*}{1.9e-06} & \multirow{2}{*}{-1.2e-05}  & k & -1.3e-03 & -1.1e-03  & q & -2.3e-04 & 2.7e-05 \\
&   &  &  &   &  &  &   & v & 3.9e-04 & 3.4e-04  & o & 1.9e-03 & 2.4e-03 \\\cmidrule(lr){2-14}
  & \multirow{2}{*}{\textbf{InsTag}}  & \multirow{2}{*}{k - v} & \multirow{2}{*}{-6.3e-04} & \multirow{2}{*}{-8.8e-05}  & \multirow{2}{*}{q - o} & \multirow{2}{*}{4.0e-06} & \multirow{2}{*}{1.3e-07}  & k & -1.2e-03 & -7.4e-04  & q & 3.0e-05 & -1.5e-05 \\
&   &  &  &   &  &  &   & v & -5.1e-04 & 6.8e-06  & o & 1.5e-03 & 1.7e-03 \\\cmidrule(lr){2-14}
  & \multirow{2}{*}{\textbf{Reward}}  & \multirow{2}{*}{k - v} & \multirow{2}{*}{-1.1e-04} & \multirow{2}{*}{-1.2e-04}  & \multirow{2}{*}{q - o} & \multirow{2}{*}{7.9e-06} & \multirow{2}{*}{4.6e-06}  & k & -6.5e-04 & -1.1e-03  & q & -2.5e-04 & -1.2e-04 \\
&   &  &  &   &  &  &   & v & -1.7e-04 & -8.0e-04  & o & 9.4e-04 & 1.4e-03 \\\bottomrule
\end{tabular}
}
\caption{Qwen2.5-3B - WizardLM - Similarity-based Metrics}
\label{tab:qwen2.5-3b_wizardlm_cosine}
\end{table*}

\begin{table*}[t]
\centering
\resizebox{\textwidth}{!}{%
\begin{tabular}{l|l|
                c c c c|
                c c c c|
                c c c c|
                c c c c}
\toprule
\multirow{2}{*}{\textbf{Dataset}} & \multirow{2}{*}{\textbf{Metrics}} 
& \multicolumn{8}{c|}{\textbf{Nuclear Norm}}
& \multicolumn{8}{c}{\textbf{Effective Rank}} \\
\cmidrule(lr){3-6}\cmidrule(lr){7-10}\cmidrule(lr){11-14}\cmidrule(lr){15-18}
 & & \textbf{Proj} & High & Low & Gap 
   & \textbf{Proj} & High & Low & Gap
   & \textbf{Proj} & High & Low & Gap
   & \textbf{Proj} & High & Low & Gap \\
\midrule
\multirow{8}{*}{\textbf{OpenHermes 2.5}}& \multirow{2}{*}{\textbf{Difficulty}}  & k & 1.3 & 3.0 & \cellcolor{red!20}{-1.8}  & q & 1.4 & 2.8 & \cellcolor{red!20}{-1.3}  & k & 64.7 & 24.1 & \cellcolor{green!20}{40.6}  & q & 115.5 & 29.2 & \cellcolor{green!20}{86.3} \\
&   & v & 2.2 & 5.6 & \cellcolor{red!20}{-3.4}  & o & 2.7 & 5.2 & \cellcolor{red!20}{-2.5}  & v & 85.8 & 28.7 & \cellcolor{green!20}{57.1}  & o & 153.7 & 32.7 & \cellcolor{green!20}{121.0} \\\cmidrule(lr){2-18}
 & \multirow{2}{*}{\textbf{IFD (GPT2)}}  & k & 1.2 & 7.5 & \cellcolor{red!20}{-6.3}  & q & 1.4 & 6.6 & \cellcolor{red!20}{-5.2}  & k & 57.0 & 25.0 & \cellcolor{green!20}{32.0}  & q & 95.4 & 28.5 & \cellcolor{green!20}{67.0} \\
&   & v & 2.3 & 13.2 & \cellcolor{red!20}{-10.9}  & o & 2.7 & 11.8 & \cellcolor{red!20}{-9.1}  & v & 74.9 & 28.3 & \cellcolor{green!20}{46.6}  & o & 121.0 & 33.2 & \cellcolor{green!20}{87.7} \\\cmidrule(lr){2-18}
 & \multirow{2}{*}{\textbf{InsTag}}  & k & 1.2 & 3.0 & \cellcolor{red!20}{-1.8}  & q & 1.4 & 2.7 & \cellcolor{red!20}{-1.3}  & k & 67.7 & 22.9 & \cellcolor{green!20}{44.8}  & q & 124.6 & 27.9 & \cellcolor{green!20}{96.7} \\
&   & v & 2.1 & 5.4 & \cellcolor{red!20}{-3.3}  & o & 2.6 & 5.0 & \cellcolor{red!20}{-2.4}  & v & 89.0 & 27.4 & \cellcolor{green!20}{61.7}  & o & 167.0 & 31.7 & \cellcolor{green!20}{135.3} \\\cmidrule(lr){2-18}
 & \multirow{2}{*}{\textbf{Reward}}  & k & 0.9 & 6.5 & \cellcolor{red!20}{-5.6}  & q & 1.1 & 5.7 & \cellcolor{red!20}{-4.6}  & k & 67.8 & 28.2 & \cellcolor{green!20}{39.6}  & q & 122.5 & 35.2 & \cellcolor{green!20}{87.2} \\
&   & v & 1.7 & 12.0 & \cellcolor{red!20}{-10.3}  & o & 2.2 & 10.7 & \cellcolor{red!20}{-8.5}  & v & 90.4 & 32.5 & \cellcolor{green!20}{57.9}  & o & 163.5 & 40.7 & \cellcolor{green!20}{122.8} \\\bottomrule
\end{tabular}
}
\caption{Qwen2.5-3B - OpenHermes 2.5 - SVD-based Metrics}
\label{tab:qwen2.5-3b_openhermes_2.5_nuclear_erank}
\end{table*}

\begin{table*}[t]
\centering
\resizebox{\textwidth}{!}{%
\begin{tabular}{l|l|
                c c c|
                c c c|
                c c c|
                c c c}
\toprule
\multirow{2}{*}{\textbf{Dataset}} & \multirow{2}{*}{\textbf{Metric}} 
& \multicolumn{6}{c|}{\textbf{Same-layer Similarity}}
& \multicolumn{6}{c}{\textbf{Adjacent-layer Similarity}} \\
\cmidrule(lr){3-5}\cmidrule(lr){6-8}\cmidrule(lr){9-11}\cmidrule(lr){12-14}
 & & \textbf{Proj} & High & Low  
   & \textbf{Proj} & High & Low
   & \textbf{Proj} & High & Low
   & \textbf{Proj} & High & Low\\
\midrule
\multirow{8}{*}{\textbf{OpenHermes 2.5}} & \multirow{2}{*}{\textbf{Difficulty}}  & \multirow{2}{*}{k - v} & \multirow{2}{*}{2.5e-04} & \multirow{2}{*}{-1.2e-04}  & \multirow{2}{*}{q - o} & \multirow{2}{*}{2.2e-08} & \multirow{2}{*}{-6.8e-06}  & k & -1.7e-03 & -1.5e-03  & q & -2.0e-04 & -9.2e-05 \\
&   &  &  &   &  &  &   & v & 9.3e-05 & -2.7e-04  & o & 8.0e-05 & 1.5e-03 \\\cmidrule(lr){2-14}
  & \multirow{2}{*}{\textbf{IFD (GPT2)}}  & \multirow{2}{*}{k - v} & \multirow{2}{*}{-9.5e-05} & \multirow{2}{*}{-5.5e-04}  & \multirow{2}{*}{q - o} & \multirow{2}{*}{2.9e-06} & \multirow{2}{*}{-6.5e-06}  & k & -1.3e-03 & 2.7e-05  & q & -2.3e-04 & -5.9e-04 \\
&   &  &  &   &  &  &   & v & -2.0e-04 & -9.0e-04  & o & 1.6e-03 & 2.1e-03 \\\cmidrule(lr){2-14}
  & \multirow{2}{*}{\textbf{InsTag}}  & \multirow{2}{*}{k - v} & \multirow{2}{*}{-7.0e-04} & \multirow{2}{*}{8.5e-05}  & \multirow{2}{*}{q - o} & \multirow{2}{*}{-4.5e-06} & \multirow{2}{*}{-3.6e-06}  & k & -1.7e-03 & -1.5e-03  & q & 1.6e-04 & 2.3e-04 \\
&   &  &  &   &  &  &   & v & -1.1e-04 & -4.9e-04  & o & 9.8e-04 & 1.7e-03 \\\cmidrule(lr){2-14}
  & \multirow{2}{*}{\textbf{Reward}}  & \multirow{2}{*}{k - v} & \multirow{2}{*}{-2.2e-04} & \multirow{2}{*}{2.8e-05}  & \multirow{2}{*}{q - o} & \multirow{2}{*}{5.4e-06} & \multirow{2}{*}{-4.2e-07}  & k & -1.7e-03 & -1.5e-03  & q & -2.1e-04 & -4.3e-04 \\
&   &  &  &   &  &  &   & v & 5.4e-04 & -1.6e-04  & o & -3.6e-05 & 1.2e-03 \\\bottomrule
\end{tabular}
}
\caption{Qwen2.5-3B - OpenHermes 2.5 - Similarity-based Metrics}
\label{tab:qwen2.5-3b_openhermes_2.5_cosine}
\end{table*}

\begin{table*}[t]
\centering
\resizebox{\textwidth}{!}{%
\begin{tabular}{l|l|
                c c c c|
                c c c c|
                c c c c|
                c c c c}
\toprule
\multirow{2}{*}{\textbf{Dataset}} & \multirow{2}{*}{\textbf{Metrics}} 
& \multicolumn{8}{c|}{\textbf{Nuclear Norm}}
& \multicolumn{8}{c}{\textbf{Effective Rank}} \\
\cmidrule(lr){3-6}\cmidrule(lr){7-10}\cmidrule(lr){11-14}\cmidrule(lr){15-18}
 & & \textbf{Proj} & High & Low & Gap 
   & \textbf{Proj} & High & Low & Gap
   & \textbf{Proj} & High & Low & Gap
   & \textbf{Proj} & High & Low & Gap \\
\midrule
\multirow{8}{*}{\textbf{Magpie}}& \multirow{2}{*}{\textbf{Difficulty}}  & k & 1.1 & 1.2 & \cellcolor{green!20}{-0.0}  & q & 1.3 & 1.3 & \cellcolor{green!20}{-0.0}  & k & 63.7 & 59.6 & \cellcolor{green!20}{4.1}  & q & 122.4 & 101.8 & \cellcolor{green!20}{20.6} \\
&   & v & 2.1 & 2.2 & \cellcolor{red!20}{-0.1}  & o & 2.5 & 2.6 & \cellcolor{red!20}{-0.1}  & v & 82.3 & 78.5 & \cellcolor{green!20}{3.9}  & o & 165.4 & 131.6 & \cellcolor{green!20}{33.9} \\\cmidrule(lr){2-18}
 & \multirow{2}{*}{\textbf{IFD (GPT2)}}  & k & 1.1 & 0.9 & \cellcolor{green!20}{0.1}  & q & 1.3 & 1.1 & \cellcolor{green!20}{0.3}  & k & 68.0 & 63.9 & \cellcolor{green!20}{4.1}  & q & 131.9 & 111.7 & \cellcolor{green!20}{20.2} \\
&   & v & 2.1 & 1.7 & \cellcolor{green!20}{0.3}  & o & 2.6 & 2.0 & \cellcolor{green!20}{0.6}  & v & 90.0 & 81.0 & \cellcolor{green!20}{9.1}  & o & 175.3 & 153.0 & \cellcolor{green!20}{22.3} \\\cmidrule(lr){2-18}
 & \multirow{2}{*}{\textbf{InsTag}}  & k & 1.2 & 1.1 & \cellcolor{green!20}{0.0}  & q & 1.4 & 1.3 & \cellcolor{green!20}{0.1}  & k & 70.6 & 59.8 & \cellcolor{green!20}{10.9}  & q & 140.0 & 106.6 & \cellcolor{green!20}{33.4} \\
&   & v & 2.1 & 2.1 & \cellcolor{green!20}{0.0}  & o & 2.6 & 2.5 & \cellcolor{green!20}{0.1}  & v & 91.3 & 78.5 & \cellcolor{green!20}{12.9}  & o & 192.1 & 139.8 & \cellcolor{green!20}{52.3} \\\cmidrule(lr){2-18}
 & \multirow{2}{*}{\textbf{Reward}}  & k & 1.1 & 1.1 & \cellcolor{green!20}{-0.0}  & q & 1.3 & 1.2 & \cellcolor{green!20}{0.1}  & k & 67.3 & 55.2 & \cellcolor{green!20}{12.1}  & q & 133.2 & 104.0 & \cellcolor{green!20}{29.3} \\
&   & v & 2.0 & 2.0 & \cellcolor{green!20}{-0.0}  & o & 2.5 & 2.3 & \cellcolor{green!20}{0.2}  & v & 90.5 & 69.6 & \cellcolor{green!20}{21.0}  & o & 181.8 & 140.0 & \cellcolor{green!20}{41.7} \\\bottomrule
\end{tabular}
}
\caption{Qwen2.5-3B - Magpie - SVD-based Metrics}
\label{tab:qwen2.5-3b_magpie_nuclear_erank}
\end{table*}

\begin{table*}[t]
\centering
\resizebox{\textwidth}{!}{%
\begin{tabular}{l|l|
                c c c|
                c c c|
                c c c|
                c c c}
\toprule
\multirow{2}{*}{\textbf{Dataset}} & \multirow{2}{*}{\textbf{Metric}} 
& \multicolumn{6}{c|}{\textbf{Same-layer Similarity}}
& \multicolumn{6}{c}{\textbf{Adjacent-layer Similarity}} \\
\cmidrule(lr){3-5}\cmidrule(lr){6-8}\cmidrule(lr){9-11}\cmidrule(lr){12-14}
 & & \textbf{Proj} & High & Low  
   & \textbf{Proj} & High & Low
   & \textbf{Proj} & High & Low
   & \textbf{Proj} & High & Low\\
\midrule
\multirow{8}{*}{\textbf{Magpie}} & \multirow{2}{*}{\textbf{Difficulty}}  & \multirow{2}{*}{k - v} & \multirow{2}{*}{-1.8e-04} & \multirow{2}{*}{-4.3e-04}  & \multirow{2}{*}{q - o} & \multirow{2}{*}{-1.1e-05} & \multirow{2}{*}{4.3e-06}  & k & -6.9e-04 & -1.6e-03  & q & -8.0e-04 & -5.8e-04 \\
&   &  &  &   &  &  &   & v & -1.3e-04 & 1.5e-04  & o & 5.2e-04 & 1.7e-03 \\\cmidrule(lr){2-14}
  & \multirow{2}{*}{\textbf{IFD (GPT2)}}  & \multirow{2}{*}{k - v} & \multirow{2}{*}{-2.8e-04} & \multirow{2}{*}{-7.9e-05}  & \multirow{2}{*}{q - o} & \multirow{2}{*}{-7.6e-06} & \multirow{2}{*}{1.5e-06}  & k & -2.1e-03 & -1.7e-03  & q & -2.7e-04 & -4.9e-04 \\
&   &  &  &   &  &  &   & v & -4.4e-04 & 3.8e-05  & o & 1.3e-03 & -9.8e-05 \\\cmidrule(lr){2-14}
  & \multirow{2}{*}{\textbf{InsTag}}  & \multirow{2}{*}{k - v} & \multirow{2}{*}{-3.9e-04} & \multirow{2}{*}{-2.8e-04}  & \multirow{2}{*}{q - o} & \multirow{2}{*}{1.1e-06} & \multirow{2}{*}{-1.0e-06}  & k & -1.8e-03 & -1.6e-03  & q & -5.0e-04 & -5.7e-04 \\
&   &  &  &   &  &  &   & v & 1.7e-04 & 2.2e-04  & o & 5.0e-04 & 1.1e-03 \\\cmidrule(lr){2-14}
  & \multirow{2}{*}{\textbf{Reward}}  & \multirow{2}{*}{k - v} & \multirow{2}{*}{-1.3e-03} & \multirow{2}{*}{3.4e-05}  & \multirow{2}{*}{q - o} & \multirow{2}{*}{-2.1e-06} & \multirow{2}{*}{-1.0e-05}  & k & -1.2e-03 & 1.8e-04  & q & -3.5e-04 & -1.7e-03 \\
&   &  &  &   &  &  &   & v & 4.8e-05 & 1.7e-03  & o & 1.8e-03 & 1.5e-03 \\\bottomrule
\end{tabular}
}
\caption{Qwen2.5-3B - Magpie - Similarity-based Metrics}
\label{tab:qwen2.5-3b_magpie_cosine}
\end{table*}

\begin{table*}[t]
\centering
\resizebox{\textwidth}{!}{%
\begin{tabular}{l|l|
                c c c c|
                c c c c|
                c c c c|
                c c c c}
\toprule
\multirow{2}{*}{\textbf{Dataset}} & \multirow{2}{*}{\textbf{Metrics}} 
& \multicolumn{8}{c|}{\textbf{Nuclear Norm}}
& \multicolumn{8}{c}{\textbf{Effective Rank}} \\
\cmidrule(lr){3-6}\cmidrule(lr){7-10}\cmidrule(lr){11-14}\cmidrule(lr){15-18}
 & & \textbf{Proj} & High & Low & Gap 
   & \textbf{Proj} & High & Low & Gap
   & \textbf{Proj} & High & Low & Gap
   & \textbf{Proj} & High & Low & Gap \\
\midrule
\multirow{2}{*}{\textbf{Reasoning}}& \multirow{2}{*}{\textbf{Reasoning}}  & k & 0.7 & 0.9 & \cellcolor{red!20}{-0.1}  & q & 1.0 & 1.1 & \cellcolor{red!20}{-0.1}  & k & 81.4 & 69.3 & \cellcolor{green!20}{12.1}  & q & 243.4 & 158.6 & \cellcolor{green!20}{84.8} \\
&   & v & 1.3 & 1.6 & \cellcolor{red!20}{-0.3}  & o & 1.9 & 2.1 & \cellcolor{red!20}{-0.2}  & v & 101.2 & 87.5 & \cellcolor{green!20}{13.8}  & o & 344.6 & 217.6 & \cellcolor{green!20}{127.0} \\
\bottomrule
\end{tabular}
}
\caption{Qwen2.5-3B - Reasoning - SVD-based Metrics}
\label{tab:qwen2.5-3b_reasoning_nuclear_erank}
\end{table*}

\begin{table*}[t]
\centering
\resizebox{\textwidth}{!}{%
\begin{tabular}{l|l|
                c c c|
                c c c|
                c c c|
                c c c}
\toprule
\multirow{2}{*}{\textbf{Dataset}} & \multirow{2}{*}{\textbf{Metric}} 
& \multicolumn{6}{c|}{\textbf{Same-layer Similarity}}
& \multicolumn{6}{c}{\textbf{Adjacent-layer Similarity}} \\
\cmidrule(lr){3-5}\cmidrule(lr){6-8}\cmidrule(lr){9-11}\cmidrule(lr){12-14}
 & & \textbf{Proj} & High & Low  
   & \textbf{Proj} & High & Low
   & \textbf{Proj} & High & Low
   & \textbf{Proj} & High & Low\\
\midrule
\multirow{2}{*}{\textbf{Reasoning}} & \multirow{2}{*}{\textbf{Reasoning}}  & \multirow{2}{*}{k - v} & \multirow{2}{*}{-3.0e-04} & \multirow{2}{*}{1.7e-04}  & \multirow{2}{*}{q - o} & \multirow{2}{*}{8.3e-07} & \multirow{2}{*}{5.4e-06}  & k & -2.7e-03 & -3.1e-03  & q & -6.9e-04 & 5.4e-05 \\
&   &  &  &   &  &  &   & v & 5.8e-04 & -2.3e-03  & o & -3.0e-03 & -1.9e-03 \\
\bottomrule
\end{tabular}
}
\caption{Qwen2.5-3B - Reasoning - Similarity-based Metrics}
\label{tab:qwen2.5-3b_reasoning_cosine}
\end{table*}

\begin{figure*}[t]
    \centering
    \includegraphics[width=1\textwidth]{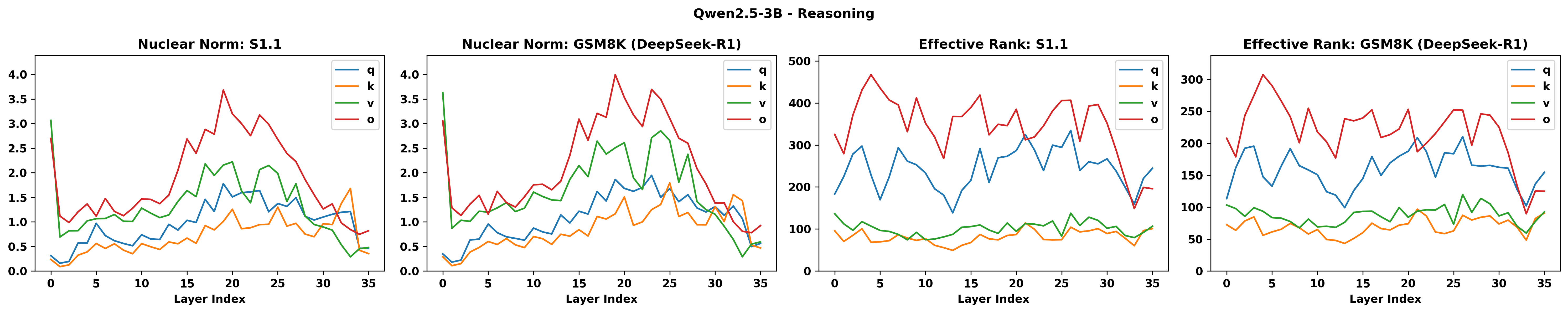}
    \caption{Qwen2.5 3B - Reasoning Data}
    \label{fig:qwen25_3b_reasoning}
\end{figure*}

\begin{figure*}[t]
    \centering
    \includegraphics[width=1\textwidth]{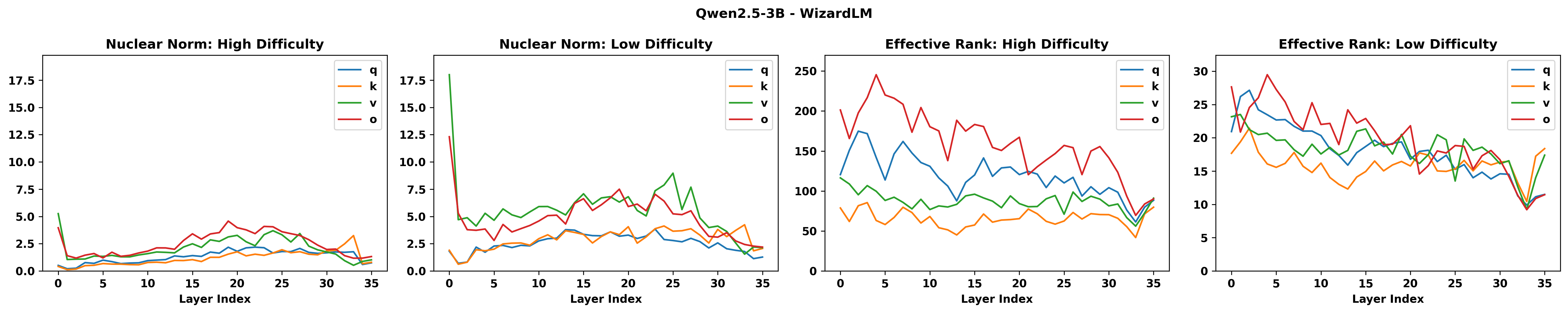}
    \caption{Qwen2.5 3B - WizardLM with Difficulty Metric}
    \label{fig:qwen25_3b_wiz_diff}
\end{figure*}

\begin{figure*}[t]
    \centering
    \includegraphics[width=1\textwidth]{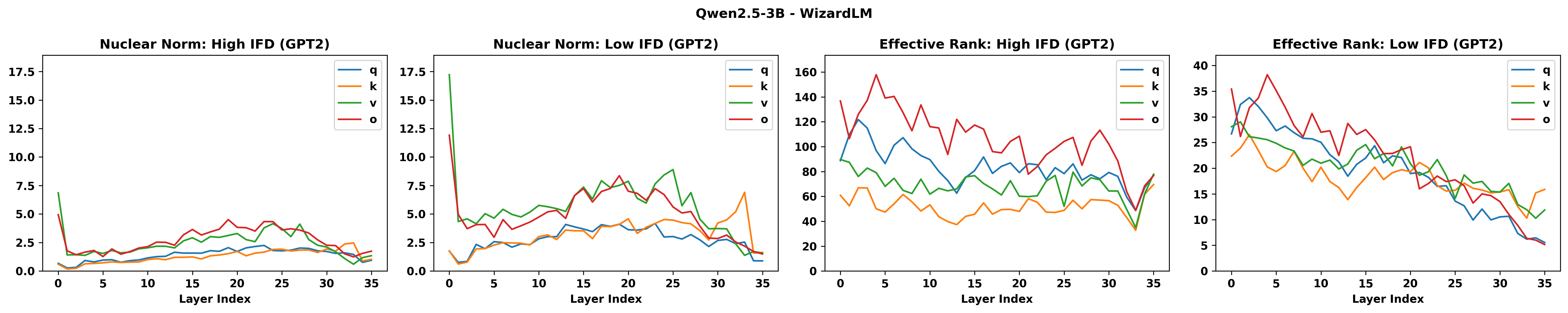}
    \caption{Qwen2.5 3B - WizardLM with IFD (GPT-2) Metric}
    \label{fig:qwen25_3b_wiz_ifd}
\end{figure*}

\begin{figure*}[t]
    \centering
    \includegraphics[width=1\textwidth]{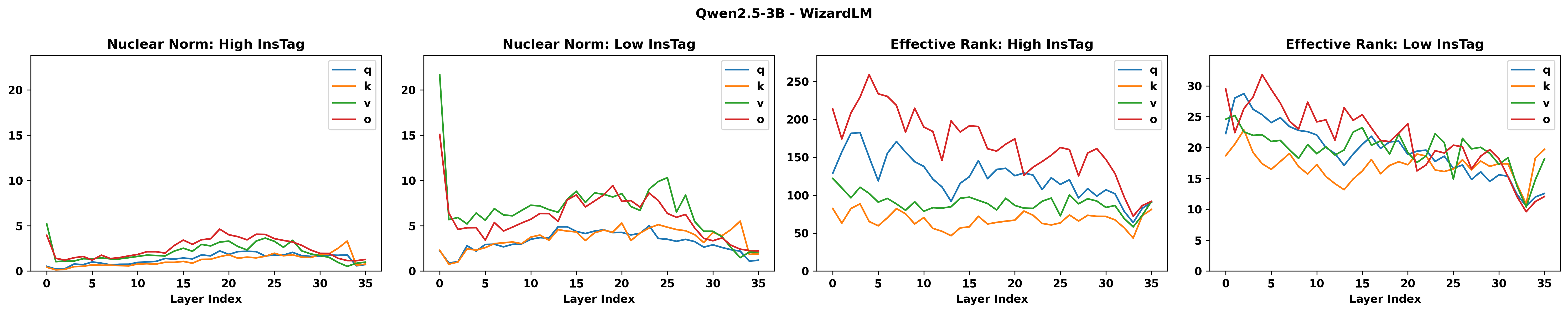}
    \caption{Qwen2.5 3B - WizardLM with InsTag Metric}
    \label{fig:qwen25_3b_wiz_instag}
\end{figure*}

\begin{figure*}[t]
    \centering
    \includegraphics[width=1\textwidth]{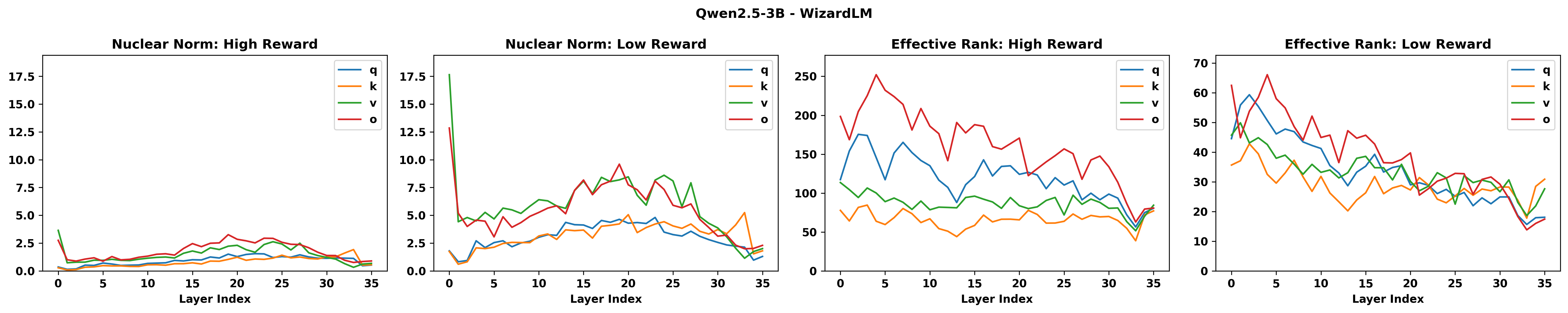}
    \caption{Qwen2.5 3B - WizardLM with Reward Model Metric}
    \label{fig:qwen25_3b_wiz_reward}
\end{figure*}

\begin{figure*}[t]
    \centering
    \includegraphics[width=1\textwidth]{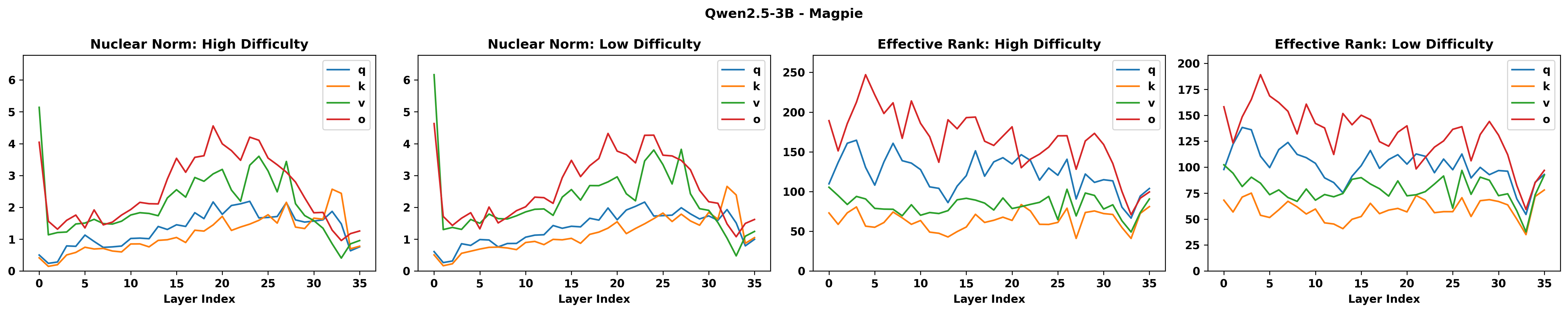}
    \caption{Qwen2.5 3B - Magpie with Difficulty Metric}
    \label{fig:qwen25_3b_mag_diff}
\end{figure*}

\begin{figure*}[t]
    \centering
    \includegraphics[width=1\textwidth]{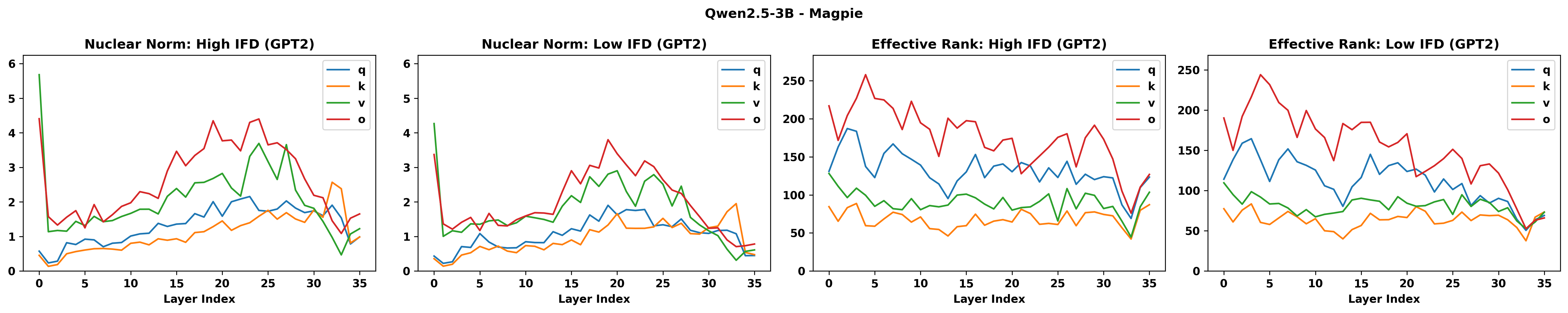}
    \caption{Qwen2.5 3B - Magpie with IFD (GPT-2) Metric}
    \label{fig:qwen25_3b_mag_ifd}
\end{figure*}

\begin{figure*}[t]
    \centering
    \includegraphics[width=1\textwidth]{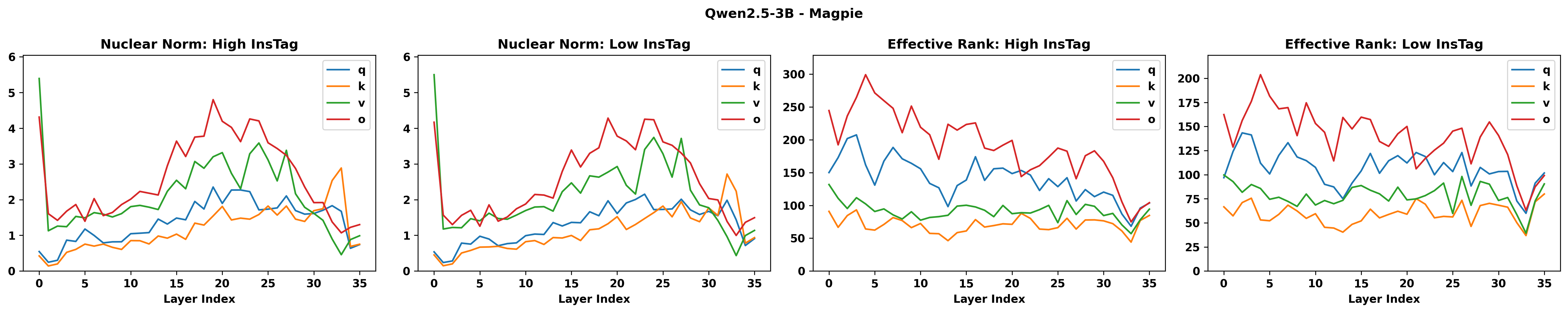}
    \caption{Qwen2.5 3B - Magpie with InsTag Metric}
    \label{fig:qwen25_3b_mag_instag}
\end{figure*}

\begin{figure*}[t]
    \centering
    \includegraphics[width=1\textwidth]{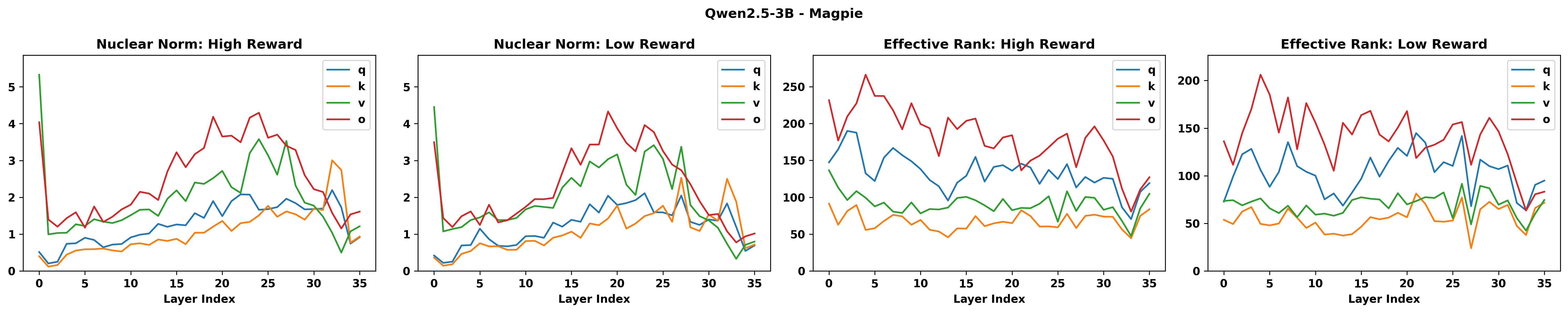}
    \caption{Qwen2.5 3B - Magpie with Reward Model Metric}
    \label{fig:qwen25_3b_mag_reward}
\end{figure*}

\begin{figure*}[t]
    \centering
    \includegraphics[width=1\textwidth]{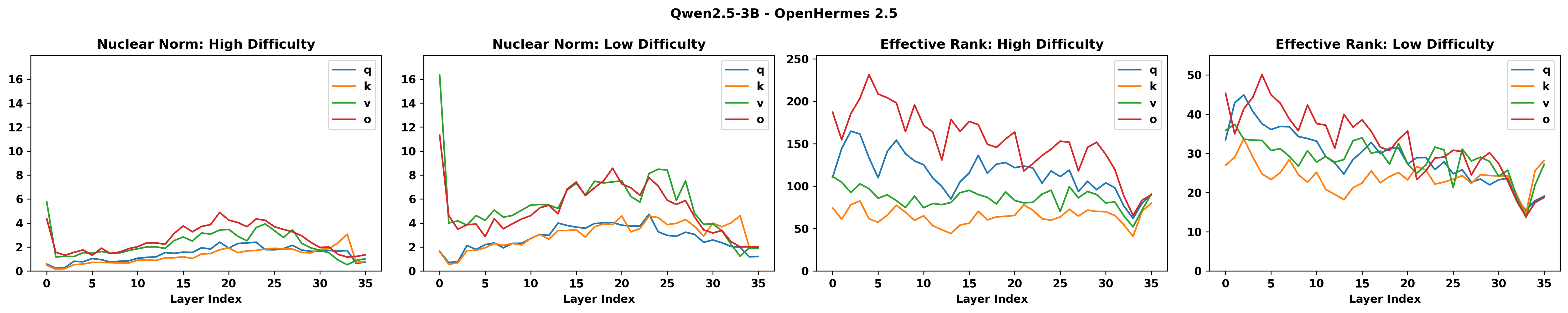}
    \caption{Qwen2.5 3B - OpenHermes with Difficulty Metric}
    \label{fig:qwen25_3b_her_diff}
\end{figure*}

\begin{figure*}[t]
    \centering
    \includegraphics[width=1\textwidth]{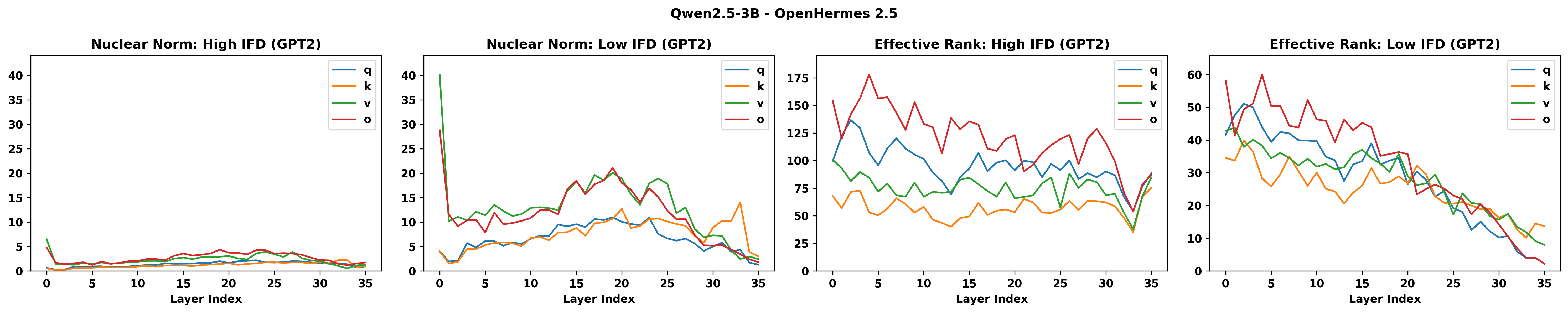}
    \caption{Qwen2.5 3B - OpenHermes with IFD (GPT-2) Metric}
    \label{fig:qwen25_3b_her_ifd}
\end{figure*}

\begin{figure*}[t]
    \centering
    \includegraphics[width=1\textwidth]{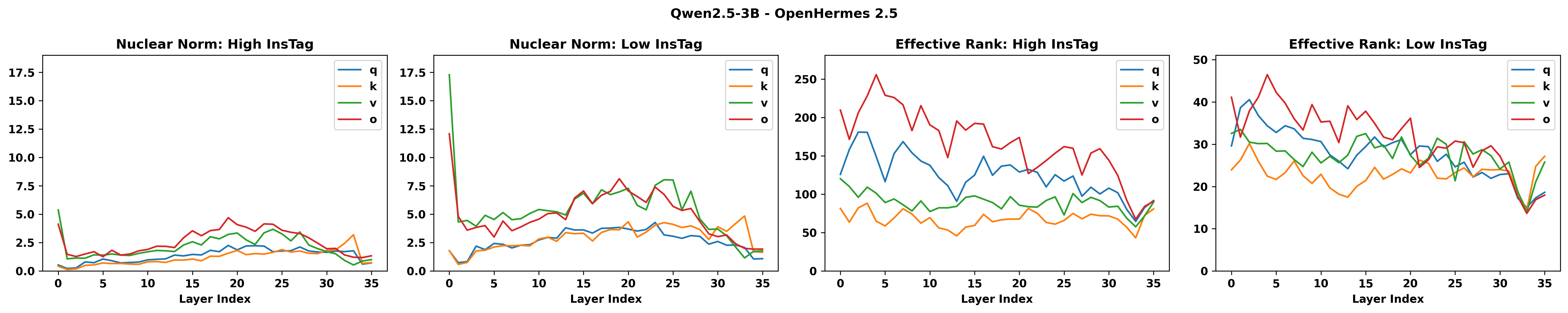}
    \caption{Qwen2.5 3B - OpenHermes with InsTag Metric}
    \label{fig:qwen25_3b_her_instag}
\end{figure*}

\begin{figure*}[t]
    \centering
    \includegraphics[width=1\textwidth]{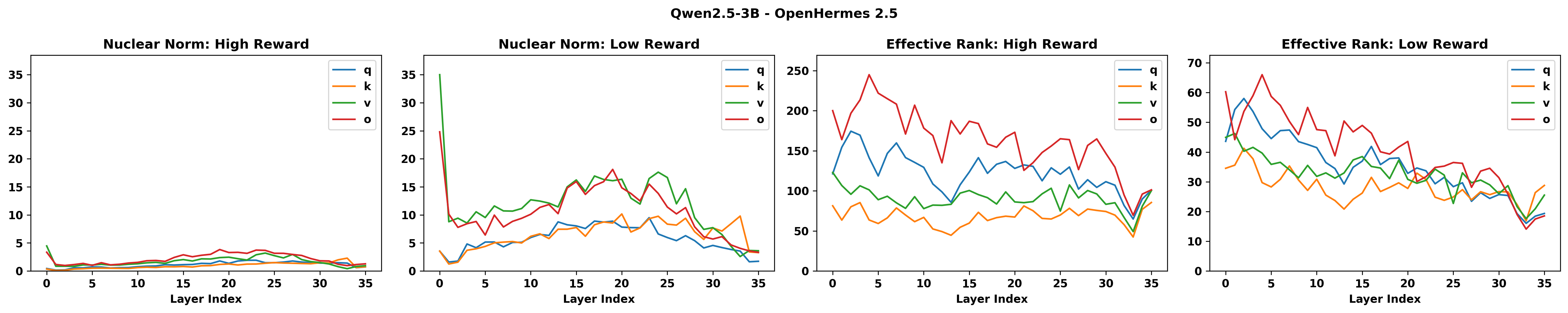}
    \caption{Qwen2.5 3B - OpenHermes with Reward Model Metric}
    \label{fig:qwen25_3b_her_reward}
\end{figure*}

\clearpage
\subsection{Qwen2.5 7B}

\begin{table*}[t]
\centering
\resizebox{\textwidth}{!}{%
\begin{tabular}{l|l|
                c c c c|
                c c c c|
                c c c c|
                c c c c}
\toprule
\multirow{2}{*}{\textbf{Dataset}} & \multirow{2}{*}{\textbf{Metrics}} 
& \multicolumn{8}{c|}{\textbf{Nuclear Norm}}
& \multicolumn{8}{c}{\textbf{Effective Rank}} \\
\cmidrule(lr){3-6}\cmidrule(lr){7-10}\cmidrule(lr){11-14}\cmidrule(lr){15-18}
 & & \textbf{Proj} & High & Low & Gap 
   & \textbf{Proj} & High & Low & Gap
   & \textbf{Proj} & High & Low & Gap
   & \textbf{Proj} & High & Low & Gap \\
\midrule
\multirow{8}{*}{\textbf{WizardLM}}& \multirow{2}{*}{\textbf{Difficulty}}  & k & 1.9 & 3.5 & \cellcolor{red!20}{-1.6}  & q & 1.9 & 2.8 & \cellcolor{red!20}{-0.9}  & k & 91.6 & 18.6 & \cellcolor{green!20}{73.0}  & q & 133.2 & 20.1 & \cellcolor{green!20}{113.2} \\
&   & v & 3.2 & 6.9 & \cellcolor{red!20}{-3.7}  & o & 3.4 & 5.2 & \cellcolor{red!20}{-1.9}  & v & 114.9 & 19.2 & \cellcolor{green!20}{95.7}  & o & 167.9 & 20.7 & \cellcolor{green!20}{147.2} \\\cmidrule(lr){2-18}
 & \multirow{2}{*}{\textbf{IFD (GPT2)}}  & k & 1.9 & 4.4 & \cellcolor{red!20}{-2.5}  & q & 1.9 & 3.3 & \cellcolor{red!20}{-1.4}  & k & 68.6 & 22.0 & \cellcolor{green!20}{46.6}  & q & 97.8 & 21.7 & \cellcolor{green!20}{76.1} \\
&   & v & 3.4 & 7.6 & \cellcolor{red!20}{-4.2}  & o & 3.5 & 5.9 & \cellcolor{red!20}{-2.5}  & v & 84.2 & 22.2 & \cellcolor{green!20}{62.0}  & o & 116.4 & 23.1 & \cellcolor{green!20}{93.3} \\\cmidrule(lr){2-18}
 & \multirow{2}{*}{\textbf{InsTag}}  & k & 1.9 & 4.1 & \cellcolor{red!20}{-2.2}  & q & 1.9 & 3.3 & \cellcolor{red!20}{-1.4}  & k & 95.6 & 20.0 & \cellcolor{green!20}{75.6}  & q & 141.2 & 21.6 & \cellcolor{green!20}{119.6} \\
&   & v & 3.2 & 7.6 & \cellcolor{red!20}{-4.5}  & o & 3.4 & 6.0 & \cellcolor{red!20}{-2.6}  & v & 120.7 & 20.8 & \cellcolor{green!20}{99.9}  & o & 180.6 & 22.4 & \cellcolor{green!20}{158.2} \\\cmidrule(lr){2-18}
 & \multirow{2}{*}{\textbf{Reward}}  & k & 1.3 & 4.3 & \cellcolor{red!20}{-3.1}  & q & 1.3 & 3.7 & \cellcolor{red!20}{-2.4}  & k & 91.5 & 35.6 & \cellcolor{green!20}{55.9}  & q & 131.5 & 38.6 & \cellcolor{green!20}{92.9} \\
&   & v & 2.2 & 7.9 & \cellcolor{red!20}{-5.7}  & o & 2.3 & 6.5 & \cellcolor{red!20}{-4.2}  & v & 113.3 & 36.9 & \cellcolor{green!20}{76.4}  & o & 166.9 & 41.5 & \cellcolor{green!20}{125.4} \\\bottomrule
\end{tabular}
}
\caption{Qwen2.5-7B - WizardLM - SVD-based Metrics}
\label{tab:qwen2.5-7b_wizardlm_nuclear_erank}
\end{table*}

\begin{table*}[t]
\centering
\resizebox{\textwidth}{!}{%
\begin{tabular}{l|l|
                c c c|
                c c c|
                c c c|
                c c c}
\toprule
\multirow{2}{*}{\textbf{Dataset}} & \multirow{2}{*}{\textbf{Metric}} 
& \multicolumn{6}{c|}{\textbf{Same-layer Similarity}}
& \multicolumn{6}{c}{\textbf{Adjacent-layer Similarity}} \\
\cmidrule(lr){3-5}\cmidrule(lr){6-8}\cmidrule(lr){9-11}\cmidrule(lr){12-14}
 & & \textbf{Proj} & High & Low  
   & \textbf{Proj} & High & Low
   & \textbf{Proj} & High & Low
   & \textbf{Proj} & High & Low\\
\midrule
\multirow{8}{*}{\textbf{WizardLM}} & \multirow{2}{*}{\textbf{Difficulty}}  & \multirow{2}{*}{k - v} & \multirow{2}{*}{-7.1e-04} & \multirow{2}{*}{-1.4e-04}  & \multirow{2}{*}{q - o} & \multirow{2}{*}{-3.9e-06} & \multirow{2}{*}{7.7e-06}  & k & -5.0e-04 & -2.4e-04  & q & -8.7e-05 & -2.3e-04 \\
&   &  &  &   &  &  &   & v & -3.2e-04 & 2.4e-04  & o & 1.1e-03 & -7.8e-05 \\\cmidrule(lr){2-14}
  & \multirow{2}{*}{\textbf{IFD (GPT2)}}  & \multirow{2}{*}{k - v} & \multirow{2}{*}{-5.1e-04} & \multirow{2}{*}{-1.8e-04}  & \multirow{2}{*}{q - o} & \multirow{2}{*}{-4.3e-06} & \multirow{2}{*}{-5.9e-06}  & k & -1.3e-03 & -6.0e-04  & q & -9.1e-06 & 2.5e-04 \\
&   &  &  &   &  &  &   & v & 1.0e-04 & -2.3e-05  & o & 7.7e-04 & -7.3e-04 \\\cmidrule(lr){2-14}
  & \multirow{2}{*}{\textbf{InsTag}}  & \multirow{2}{*}{k - v} & \multirow{2}{*}{-1.1e-03} & \multirow{2}{*}{-5.0e-05}  & \multirow{2}{*}{q - o} & \multirow{2}{*}{8.8e-07} & \multirow{2}{*}{1.5e-06}  & k & -5.2e-04 & -7.3e-04  & q & -4.4e-05 & 1.6e-04 \\
&   &  &  &   &  &  &   & v & 1.3e-04 & 4.6e-04  & o & 1.2e-03 & -1.6e-04 \\\cmidrule(lr){2-14}
  & \multirow{2}{*}{\textbf{Reward}}  & \multirow{2}{*}{k - v} & \multirow{2}{*}{-6.2e-04} & \multirow{2}{*}{-3.8e-04}  & \multirow{2}{*}{q - o} & \multirow{2}{*}{2.3e-06} & \multirow{2}{*}{1.3e-06}  & k & -5.8e-04 & -7.5e-04  & q & -1.3e-04 & 1.9e-04 \\
&   &  &  &   &  &  &   & v & 1.7e-04 & 9.3e-05  & o & 5.5e-04 & -4.5e-04 \\\bottomrule
\end{tabular}
}
\caption{Qwen2.5-7B - WizardLM - Similarity-based Metrics}
\label{tab:qwen2.5-7b_wizardlm_cosine}
\end{table*}

\begin{table*}[t]
\centering
\resizebox{\textwidth}{!}{%
\begin{tabular}{l|l|
                c c c c|
                c c c c|
                c c c c|
                c c c c}
\toprule
\multirow{2}{*}{\textbf{Dataset}} & \multirow{2}{*}{\textbf{Metrics}} 
& \multicolumn{8}{c|}{\textbf{Nuclear Norm}}
& \multicolumn{8}{c}{\textbf{Effective Rank}} \\
\cmidrule(lr){3-6}\cmidrule(lr){7-10}\cmidrule(lr){11-14}\cmidrule(lr){15-18}
 & & \textbf{Proj} & High & Low & Gap 
   & \textbf{Proj} & High & Low & Gap
   & \textbf{Proj} & High & Low & Gap
   & \textbf{Proj} & High & Low & Gap \\
\midrule
\multirow{8}{*}{\textbf{OpenHermes 2.5}}& \multirow{2}{*}{\textbf{Difficulty}}  & k & 2.0 & 4.0 & \cellcolor{red!20}{-2.0}  & q & 2.0 & 3.3 & \cellcolor{red!20}{-1.2}  & k & 92.1 & 29.0 & \cellcolor{green!20}{63.1}  & q & 135.7 & 33.3 & \cellcolor{green!20}{102.4} \\
&   & v & 3.4 & 7.2 & \cellcolor{red!20}{-3.8}  & o & 3.6 & 5.9 & \cellcolor{red!20}{-2.3}  & v & 115.7 & 31.1 & \cellcolor{green!20}{84.6}  & o & 170.8 & 35.1 & \cellcolor{green!20}{135.7} \\\cmidrule(lr){2-18}
 & \multirow{2}{*}{\textbf{IFD (GPT2)}}  & k & 1.8 & 8.5 & \cellcolor{red!20}{-6.7}  & q & 1.9 & 6.5 & \cellcolor{red!20}{-4.7}  & k & 78.5 & 32.7 & \cellcolor{green!20}{45.8}  & q & 114.0 & 33.7 & \cellcolor{green!20}{80.3} \\
&   & v & 3.3 & 14.8 & \cellcolor{red!20}{-11.5}  & o & 3.5 & 11.8 & \cellcolor{red!20}{-8.3}  & v & 96.6 & 32.5 & \cellcolor{green!20}{64.0}  & o & 136.6 & 36.5 & \cellcolor{green!20}{100.1} \\\cmidrule(lr){2-18}
 & \multirow{2}{*}{\textbf{InsTag}}  & k & 1.8 & 4.1 & \cellcolor{red!20}{-2.3}  & q & 1.9 & 3.3 & \cellcolor{red!20}{-1.5}  & k & 98.4 & 27.4 & \cellcolor{green!20}{70.9}  & q & 145.9 & 31.5 & \cellcolor{green!20}{114.4} \\
&   & v & 3.1 & 7.2 & \cellcolor{red!20}{-4.1}  & o & 3.3 & 6.0 & \cellcolor{red!20}{-2.7}  & v & 123.1 & 29.7 & \cellcolor{green!20}{93.5}  & o & 183.8 & 33.7 & \cellcolor{green!20}{150.1} \\\cmidrule(lr){2-18}
 & \multirow{2}{*}{\textbf{Reward}}  & k & 1.5 & 9.5 & \cellcolor{red!20}{-8.1}  & q & 1.6 & 7.4 & \cellcolor{red!20}{-5.8}  & k & 98.0 & 34.4 & \cellcolor{green!20}{63.5}  & q & 143.9 & 38.5 & \cellcolor{green!20}{105.4} \\
&   & v & 2.6 & 17.2 & \cellcolor{red!20}{-14.6}  & o & 2.9 & 13.6 & \cellcolor{red!20}{-10.6}  & v & 123.4 & 36.3 & \cellcolor{green!20}{87.1}  & o & 182.6 & 41.9 & \cellcolor{green!20}{140.7} \\\bottomrule
\end{tabular}
}
\caption{Qwen2.5-7B - OpenHermes 2.5 - SVD-based Metrics}
\label{tab:qwen2.5-7b_openhermes_2.5_nuclear_erank}
\end{table*}

\begin{table*}[t]
\centering
\resizebox{\textwidth}{!}{%
\begin{tabular}{l|l|
                c c c|
                c c c|
                c c c|
                c c c}
\toprule
\multirow{2}{*}{\textbf{Dataset}} & \multirow{2}{*}{\textbf{Metric}} 
& \multicolumn{6}{c|}{\textbf{Same-layer Similarity}}
& \multicolumn{6}{c}{\textbf{Adjacent-layer Similarity}} \\
\cmidrule(lr){3-5}\cmidrule(lr){6-8}\cmidrule(lr){9-11}\cmidrule(lr){12-14}
 & & \textbf{Proj} & High & Low  
   & \textbf{Proj} & High & Low
   & \textbf{Proj} & High & Low
   & \textbf{Proj} & High & Low\\
\midrule
\multirow{8}{*}{\textbf{OpenHermes 2.5}} & \multirow{2}{*}{\textbf{Difficulty}}  & \multirow{2}{*}{k - v} & \multirow{2}{*}{-8.8e-04} & \multirow{2}{*}{-2.7e-04}  & \multirow{2}{*}{q - o} & \multirow{2}{*}{2.0e-06} & \multirow{2}{*}{-1.6e-06}  & k & -6.2e-04 & -5.6e-04  & q & -8.3e-05 & 7.2e-05 \\
&   &  &  &   &  &  &   & v & -2.3e-04 & 2.6e-04  & o & 7.1e-04 & -1.0e-04 \\\cmidrule(lr){2-14}
  & \multirow{2}{*}{\textbf{IFD (GPT2)}}  & \multirow{2}{*}{k - v} & \multirow{2}{*}{-4.0e-04} & \multirow{2}{*}{-3.0e-04}  & \multirow{2}{*}{q - o} & \multirow{2}{*}{-2.4e-06} & \multirow{2}{*}{7.5e-06}  & k & -5.3e-04 & -5.9e-04  & q & -1.5e-04 & -1.8e-05 \\
&   &  &  &   &  &  &   & v & -2.9e-04 & 4.4e-04  & o & 3.6e-04 & -1.6e-03 \\\cmidrule(lr){2-14}
  & \multirow{2}{*}{\textbf{InsTag}}  & \multirow{2}{*}{k - v} & \multirow{2}{*}{-1.1e-03} & \multirow{2}{*}{-3.1e-04}  & \multirow{2}{*}{q - o} & \multirow{2}{*}{-2.7e-06} & \multirow{2}{*}{-3.0e-06}  & k & -7.9e-04 & -3.7e-04  & q & -3.2e-05 & 2.1e-04 \\
&   &  &  &   &  &  &   & v & 1.5e-05 & 3.0e-04  & o & 1.1e-03 & -6.0e-04 \\\cmidrule(lr){2-14}
  & \multirow{2}{*}{\textbf{Reward}}  & \multirow{2}{*}{k - v} & \multirow{2}{*}{-5.4e-04} & \multirow{2}{*}{7.9e-05}  & \multirow{2}{*}{q - o} & \multirow{2}{*}{-2.1e-06} & \multirow{2}{*}{2.9e-06}  & k & -5.9e-04 & -3.3e-04  & q & -2.3e-04 & -1.2e-04 \\
&   &  &  &   &  &  &   & v & 2.2e-04 & -5.2e-05  & o & -3.6e-05 & -9.4e-05 \\\bottomrule
\end{tabular}
}
\caption{Qwen2.5-7B - OpenHermes 2.5 - Similarity-based Metrics}
\label{tab:qwen2.5-7b_openhermes_2.5_cosine}
\end{table*}

\begin{table*}[t]
\centering
\resizebox{\textwidth}{!}{%
\begin{tabular}{l|l|
                c c c c|
                c c c c|
                c c c c|
                c c c c}
\toprule
\multirow{2}{*}{\textbf{Dataset}} & \multirow{2}{*}{\textbf{Metrics}} 
& \multicolumn{8}{c|}{\textbf{Nuclear Norm}}
& \multicolumn{8}{c}{\textbf{Effective Rank}} \\
\cmidrule(lr){3-6}\cmidrule(lr){7-10}\cmidrule(lr){11-14}\cmidrule(lr){15-18}
 & & \textbf{Proj} & High & Low & Gap 
   & \textbf{Proj} & High & Low & Gap
   & \textbf{Proj} & High & Low & Gap
   & \textbf{Proj} & High & Low & Gap \\
\midrule
\multirow{8}{*}{\textbf{Magpie}}& \multirow{2}{*}{\textbf{Difficulty}}  & k & 1.8 & 1.8 & \cellcolor{green!20}{0.0}  & q & 1.9 & 1.9 & \cellcolor{green!20}{0.0}  & k & 95.9 & 83.9 & \cellcolor{green!20}{12.1}  & q & 153.3 & 124.4 & \cellcolor{green!20}{28.9} \\
&   & v & 3.1 & 3.1 & \cellcolor{red!20}{-0.1}  & o & 3.4 & 3.4 & \cellcolor{green!20}{-0.0}  & v & 118.1 & 102.8 & \cellcolor{green!20}{15.3}  & o & 195.2 & 151.8 & \cellcolor{green!20}{43.4} \\\cmidrule(lr){2-18}
 & \multirow{2}{*}{\textbf{IFD (GPT2)}}  & k & 1.8 & 1.6 & \cellcolor{green!20}{0.2}  & q & 1.9 & 1.6 & \cellcolor{green!20}{0.3}  & k & 100.6 & 93.5 & \cellcolor{green!20}{7.1}  & q & 163.7 & 133.3 & \cellcolor{green!20}{30.4} \\
&   & v & 3.1 & 2.8 & \cellcolor{green!20}{0.3}  & o & 3.5 & 2.8 & \cellcolor{green!20}{0.7}  & v & 125.8 & 110.9 & \cellcolor{green!20}{14.9}  & o & 207.2 & 170.2 & \cellcolor{green!20}{37.0} \\\cmidrule(lr){2-18}
 & \multirow{2}{*}{\textbf{InsTag}}  & k & 2.0 & 1.7 & \cellcolor{green!20}{0.3}  & q & 2.2 & 1.8 & \cellcolor{green!20}{0.3}  & k & 108.4 & 86.0 & \cellcolor{green!20}{22.4}  & q & 174.1 & 132.0 & \cellcolor{green!20}{42.1} \\
&   & v & 3.5 & 3.0 & \cellcolor{green!20}{0.4}  & o & 3.8 & 3.3 & \cellcolor{green!20}{0.5}  & v & 133.8 & 105.7 & \cellcolor{green!20}{28.1}  & o & 225.0 & 163.3 & \cellcolor{green!20}{61.7} \\\cmidrule(lr){2-18}
 & \multirow{2}{*}{\textbf{Reward}}  & k & 1.8 & 1.8 & \cellcolor{green!20}{0.0}  & q & 1.9 & 1.8 & \cellcolor{green!20}{0.1}  & k & 99.4 & 83.9 & \cellcolor{green!20}{15.5}  & q & 163.5 & 131.2 & \cellcolor{green!20}{32.3} \\
&   & v & 3.0 & 3.0 & \cellcolor{green!20}{-0.0}  & o & 3.4 & 3.2 & \cellcolor{green!20}{0.2}  & v & 126.6 & 101.1 & \cellcolor{green!20}{25.5}  & o & 212.0 & 164.1 & \cellcolor{green!20}{47.9} \\\bottomrule
\end{tabular}
}
\caption{Qwen2.5-7B - Magpie - SVD-based Metrics}
\label{tab:qwen2.5-7b_magpie_nuclear_erank}
\end{table*}

\begin{table*}[t]
\centering
\resizebox{\textwidth}{!}{%
\begin{tabular}{l|l|
                c c c|
                c c c|
                c c c|
                c c c}
\toprule
\multirow{2}{*}{\textbf{Dataset}} & \multirow{2}{*}{\textbf{Metric}} 
& \multicolumn{6}{c|}{\textbf{Same-layer Similarity}}
& \multicolumn{6}{c}{\textbf{Adjacent-layer Similarity}} \\
\cmidrule(lr){3-5}\cmidrule(lr){6-8}\cmidrule(lr){9-11}\cmidrule(lr){12-14}
 & & \textbf{Proj} & High & Low  
   & \textbf{Proj} & High & Low
   & \textbf{Proj} & High & Low
   & \textbf{Proj} & High & Low\\
\midrule
\multirow{8}{*}{\textbf{Magpie}} & \multirow{2}{*}{\textbf{Difficulty}}  & \multirow{2}{*}{k - v} & \multirow{2}{*}{-7.2e-04} & \multirow{2}{*}{-6.9e-04}  & \multirow{2}{*}{q - o} & \multirow{2}{*}{-1.6e-06} & \multirow{2}{*}{7.9e-06}  & k & -1.2e-04 & -7.1e-05  & q & 1.0e-04 & -1.0e-04 \\
&   &  &  &   &  &  &   & v & 3.1e-04 & -1.0e-04  & o & 4.1e-04 & 2.5e-04 \\\cmidrule(lr){2-14}
  & \multirow{2}{*}{\textbf{IFD (GPT2)}}  & \multirow{2}{*}{k - v} & \multirow{2}{*}{-5.9e-04} & \multirow{2}{*}{-8.9e-04}  & \multirow{2}{*}{q - o} & \multirow{2}{*}{4.6e-06} & \multirow{2}{*}{6.9e-06}  & k & -7.9e-06 & -7.9e-04  & q & -2.7e-04 & 9.7e-05 \\
&   &  &  &   &  &  &   & v & -1.1e-04 & -1.8e-04  & o & 8.6e-04 & 1.1e-03 \\\cmidrule(lr){2-14}
  & \multirow{2}{*}{\textbf{InsTag}}  & \multirow{2}{*}{k - v} & \multirow{2}{*}{-1.1e-03} & \multirow{2}{*}{-8.3e-04}  & \multirow{2}{*}{q - o} & \multirow{2}{*}{4.1e-06} & \multirow{2}{*}{2.1e-06}  & k & -3.3e-04 & -1.6e-04  & q & -9.4e-05 & 1.2e-05 \\
&   &  &  &   &  &  &   & v & 5.3e-04 & -2.7e-04  & o & 1.5e-03 & 2.7e-04 \\\cmidrule(lr){2-14}
  & \multirow{2}{*}{\textbf{Reward}}  & \multirow{2}{*}{k - v} & \multirow{2}{*}{-1.5e-03} & \multirow{2}{*}{-1.2e-03}  & \multirow{2}{*}{q - o} & \multirow{2}{*}{4.1e-06} & \multirow{2}{*}{3.9e-06}  & k & -6.9e-04 & -2.8e-04  & q & -1.8e-04 & 2.0e-05 \\
&   &  &  &   &  &  &   & v & -2.9e-04 & 6.5e-04  & o & 1.3e-03 & -1.6e-04 \\\bottomrule
\end{tabular}
}
\caption{Qwen2.5-7B - Magpie - Similarity-based Metrics}
\label{tab:qwen2.5-7b_magpie_cosine}
\end{table*}

\begin{table*}[t]
\centering
\resizebox{\textwidth}{!}{%
\begin{tabular}{l|l|
                c c c c|
                c c c c|
                c c c c|
                c c c c}
\toprule
\multirow{2}{*}{\textbf{Dataset}} & \multirow{2}{*}{\textbf{Metrics}} 
& \multicolumn{8}{c|}{\textbf{Nuclear Norm}}
& \multicolumn{8}{c}{\textbf{Effective Rank}} \\
\cmidrule(lr){3-6}\cmidrule(lr){7-10}\cmidrule(lr){11-14}\cmidrule(lr){15-18}
 & & \textbf{Proj} & High & Low & Gap 
   & \textbf{Proj} & High & Low & Gap
   & \textbf{Proj} & High & Low & Gap
   & \textbf{Proj} & High & Low & Gap \\
\midrule
\multirow{2}{*}{\textbf{Reasoning}}& \multirow{2}{*}{\textbf{Reasoning}}  & k & 1.1 & 1.3 & \cellcolor{red!20}{-0.3}  & q & 1.4 & 1.5 & \cellcolor{red!20}{-0.2}  & k & 138.8 & 106.2 & \cellcolor{green!20}{32.7}  & q & 361.3 & 203.4 & \cellcolor{green!20}{157.9} \\
&   & v & 1.9 & 2.5 & \cellcolor{red!20}{-0.6}  & o & 2.5 & 2.8 & \cellcolor{red!20}{-0.3}  & v & 170.5 & 126.7 & \cellcolor{green!20}{43.8}  & o & 509.9 & 263.1 & \cellcolor{green!20}{246.8} \\
\bottomrule
\end{tabular}
}
\caption{Qwen2.5-7B - Reasoning - SVD-based Metrics}
\label{tab:qwen2.5-7b_reasoning_nuclear_erank}
\end{table*}

\begin{table*}[t]
\centering
\resizebox{\textwidth}{!}{%
\begin{tabular}{l|l|
                c c c|
                c c c|
                c c c|
                c c c}
\toprule
\multirow{2}{*}{\textbf{Dataset}} & \multirow{2}{*}{\textbf{Metric}} 
& \multicolumn{6}{c|}{\textbf{Same-layer Similarity}}
& \multicolumn{6}{c}{\textbf{Adjacent-layer Similarity}} \\
\cmidrule(lr){3-5}\cmidrule(lr){6-8}\cmidrule(lr){9-11}\cmidrule(lr){12-14}
 & & \textbf{Proj} & High & Low  
   & \textbf{Proj} & High & Low
   & \textbf{Proj} & High & Low
   & \textbf{Proj} & High & Low\\
\midrule
\multirow{2}{*}{\textbf{Reasoning}} & \multirow{2}{*}{\textbf{Reasoning}}  & \multirow{2}{*}{k - v} & \multirow{2}{*}{-8.9e-04} & \multirow{2}{*}{8.6e-05}  & \multirow{2}{*}{q - o} & \multirow{2}{*}{-2.4e-06} & \multirow{2}{*}{-1.1e-05}  & k & -2.4e-03 & -2.7e-03  & q & 4.4e-05 & 1.7e-04 \\
&   &  &  &   &  &  &   & v & 9.7e-04 & 2.2e-03  & o & 1.6e-03 & 2.5e-03 \\
\bottomrule
\end{tabular}
}
\caption{Qwen2.5-7B - Reasoning - Similarity-based Metrics}
\label{tab:qwen2.5-7b_reasoning_cosine}
\end{table*}

\begin{figure*}[t]
    \centering
    \includegraphics[width=1\textwidth]{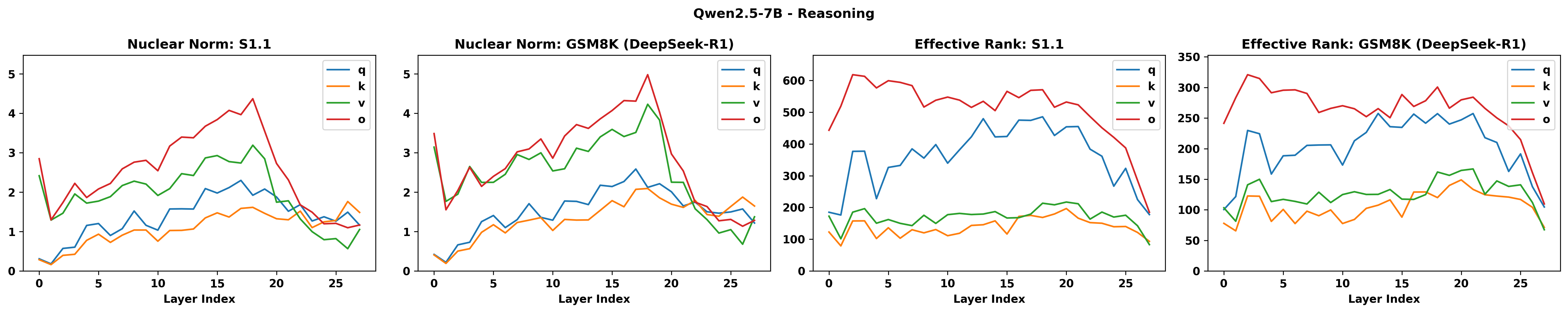}
    \caption{Qwen2.5 7B - Reasoning Data}
    \label{fig:qwen25_7b_reasoning}
\end{figure*}

\begin{figure*}[t]
    \centering
    \includegraphics[width=1\textwidth]{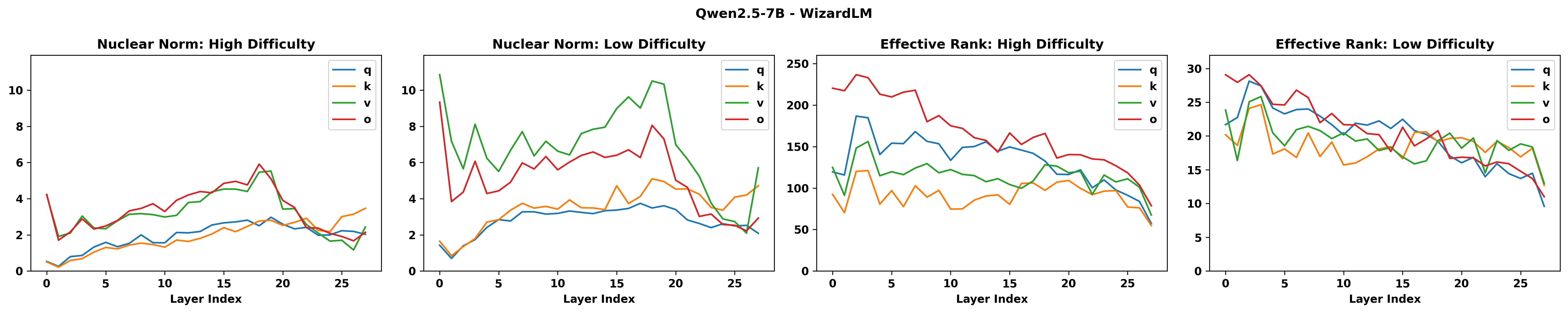}
    \caption{Qwen2.5 7B - WizardLM with Difficulty Metric}
    \label{fig:qwen25_7b_wiz_diff}
\end{figure*}

\begin{figure*}[t]
    \centering
    \includegraphics[width=1\textwidth]{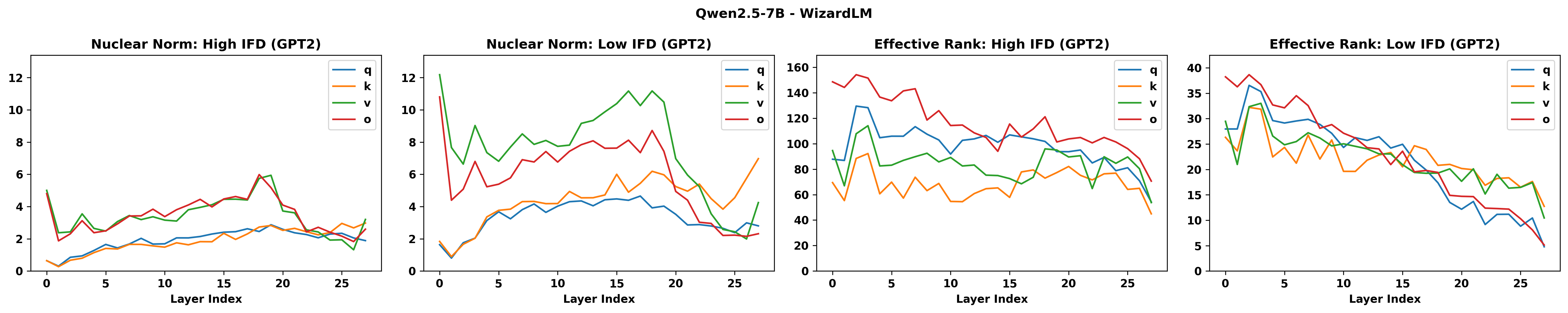}
    \caption{Qwen2.5 7B - WizardLM with IFD (GPT-2) Metric}
    \label{fig:qwen25_7b_wiz_ifd}
\end{figure*}

\begin{figure*}[t]
    \centering
    \includegraphics[width=1\textwidth]{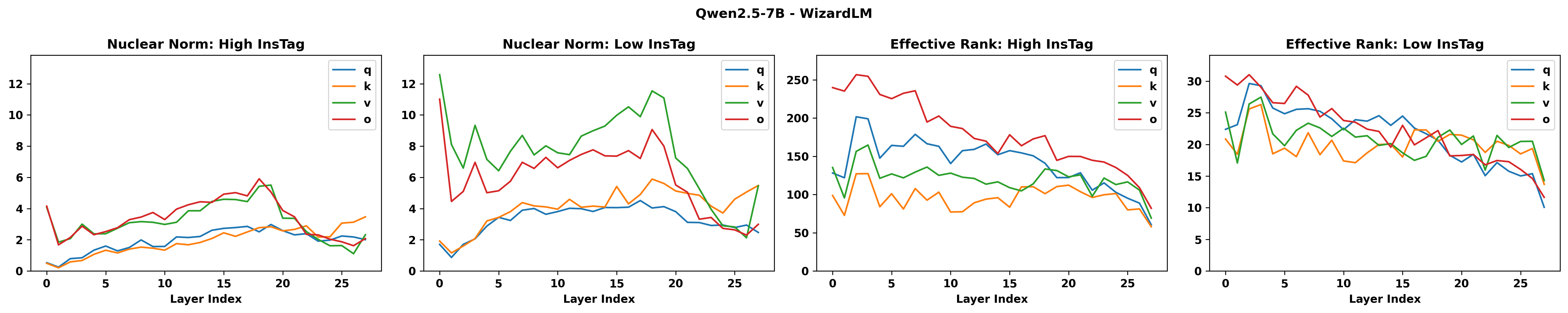}
    \caption{Qwen2.5 7B - WizardLM with InsTag Metric}
    \label{fig:qwen25_7b_wiz_instag}
\end{figure*}

\begin{figure*}[t]
    \centering
    \includegraphics[width=1\textwidth]{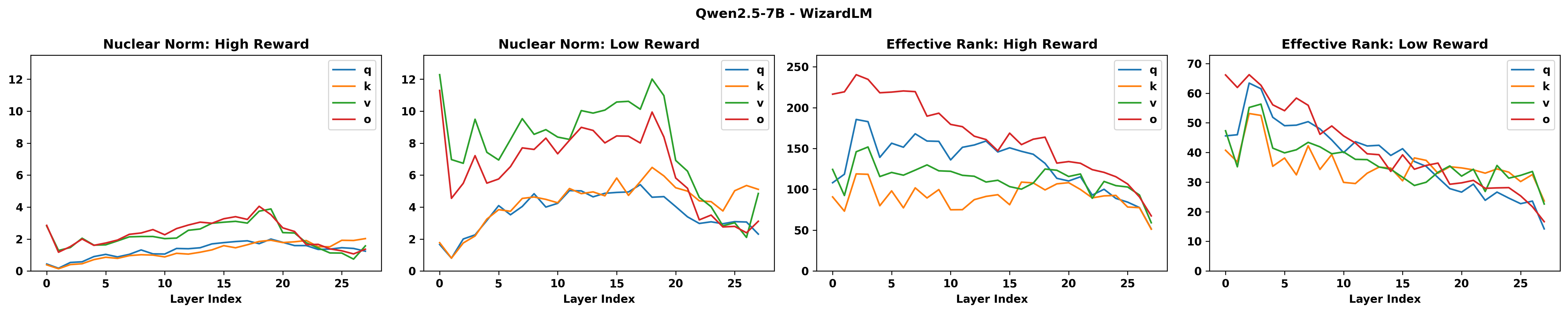}
    \caption{Qwen2.5 7B - WizardLM with Reward Model Metric}
    \label{fig:qwen25_7b_wiz_reward}
\end{figure*}

\begin{figure*}[t]
    \centering
    \includegraphics[width=1\textwidth]{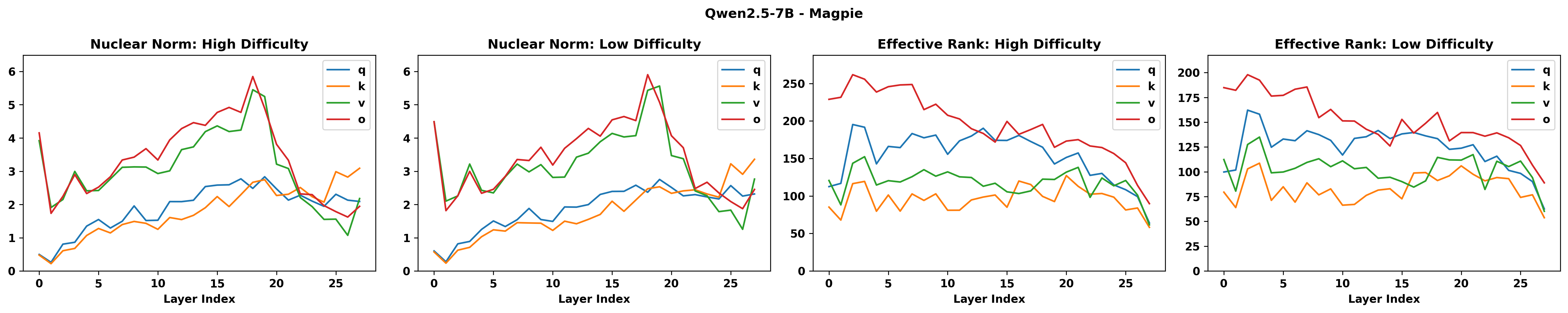}
    \caption{Qwen2.5 7B - Magpie with Difficulty Metric}
    \label{fig:qwen25_7b_mag_diff}
\end{figure*}

\begin{figure*}[t]
    \centering
    \includegraphics[width=1\textwidth]{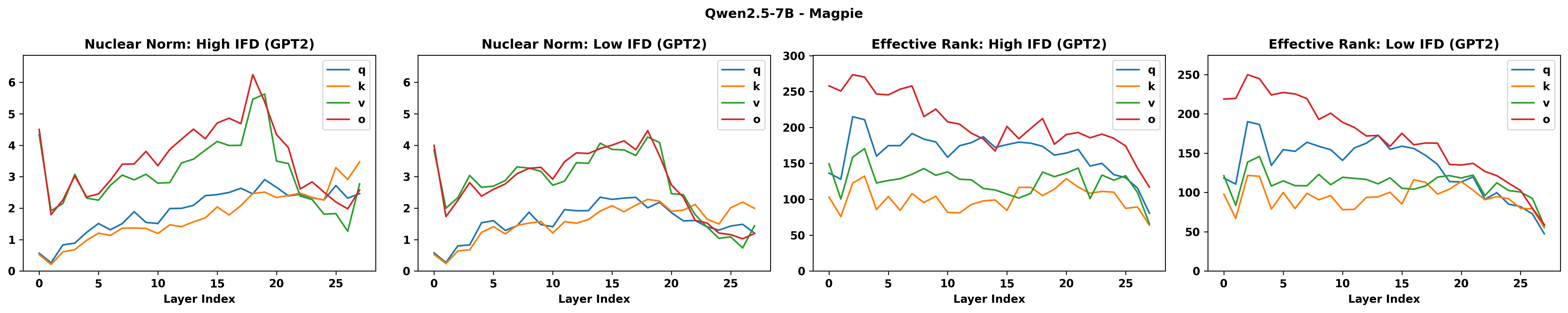}
    \caption{Qwen2.5 7B - Magpie with IFD (GPT-2) Metric}
    \label{fig:qwen25_7b_mag_ifd}
\end{figure*}

\begin{figure*}[t]
    \centering
    \includegraphics[width=1\textwidth]{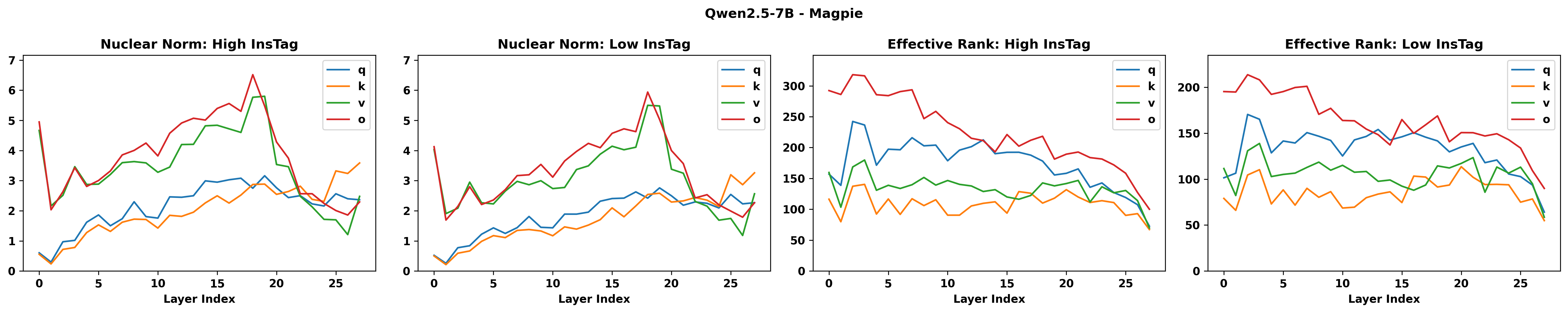}
    \caption{Qwen2.5 7B - Magpie with InsTag Metric}
    \label{fig:qwen25_7b_mag_instag}
\end{figure*}

\begin{figure*}[t]
    \centering
    \includegraphics[width=1\textwidth]{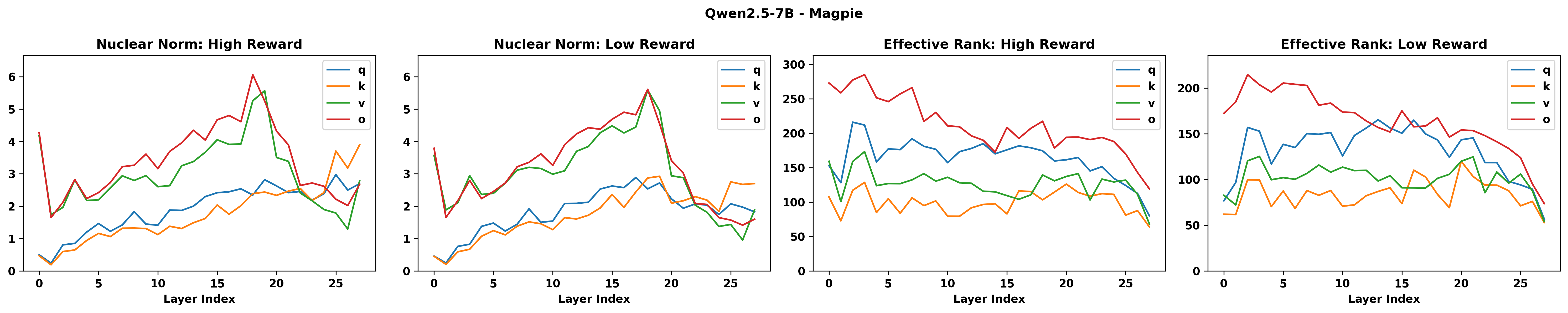}
    \caption{Qwen2.5 7B - Magpie with Reward Model Metric}
    \label{fig:qwen25_7b_mag_reward}
\end{figure*}

\begin{figure*}[t]
    \centering
    \includegraphics[width=1\textwidth]{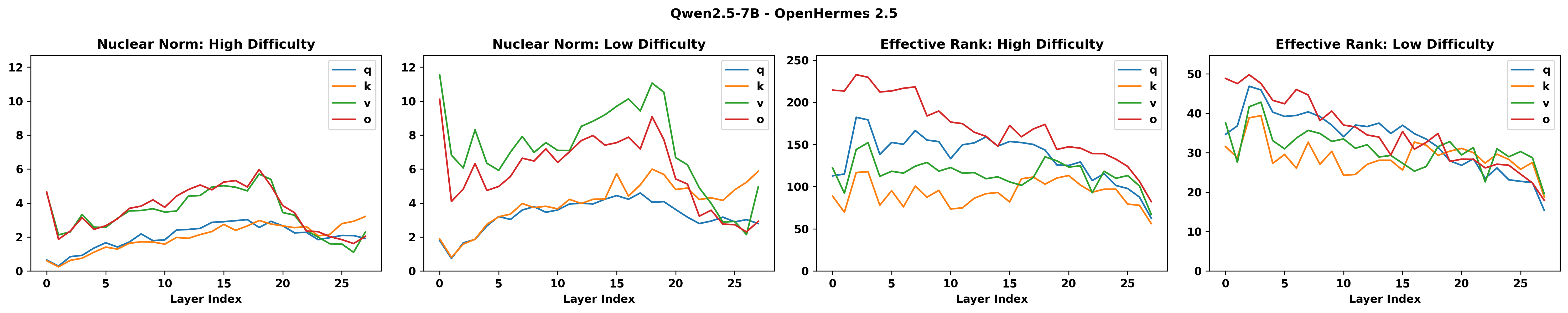}
    \caption{Qwen2.5 7B - OpenHermes with Difficulty Metric}
    \label{fig:qwen25_7b_her_diff}
\end{figure*}

\begin{figure*}[t]
    \centering
    \includegraphics[width=1\textwidth]{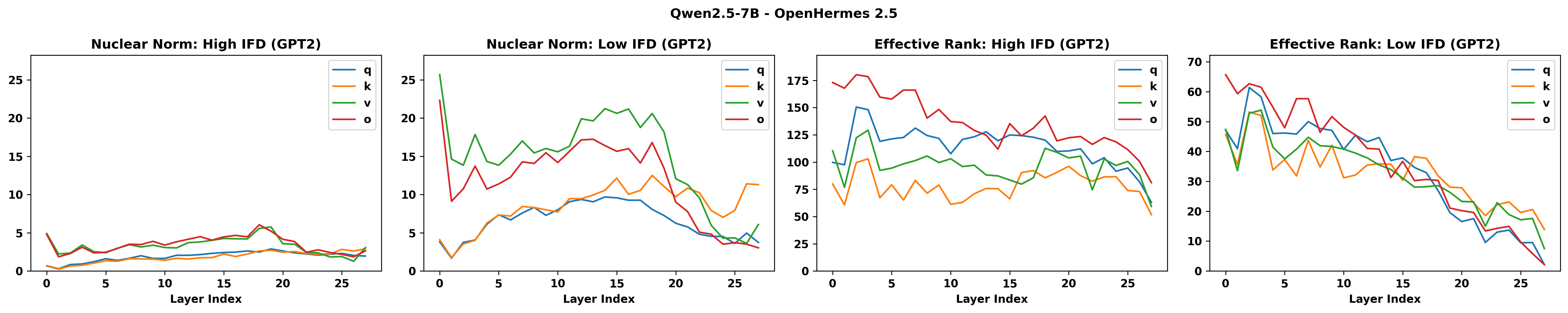}
    \caption{Qwen2.5 7B - OpenHermes with IFD (GPT-2) Metric}
    \label{fig:qwen25_7b_her_ifd}
\end{figure*}

\begin{figure*}[t]
    \centering
    \includegraphics[width=1\textwidth]{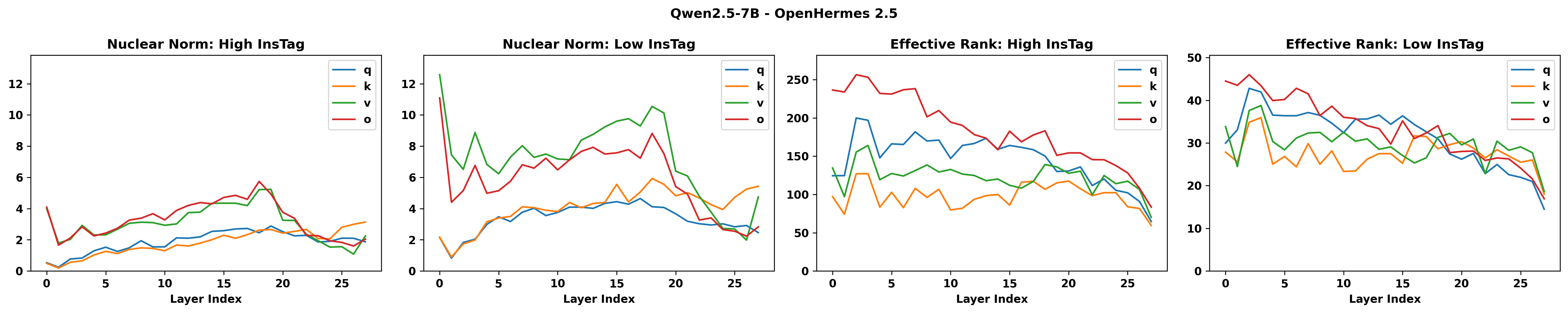}
    \caption{Qwen2.5 7B - OpenHermes with InsTag Metric}
    \label{fig:qwen25_7b_her_instag}
\end{figure*}

\begin{figure*}[t]
    \centering
    \includegraphics[width=1\textwidth]{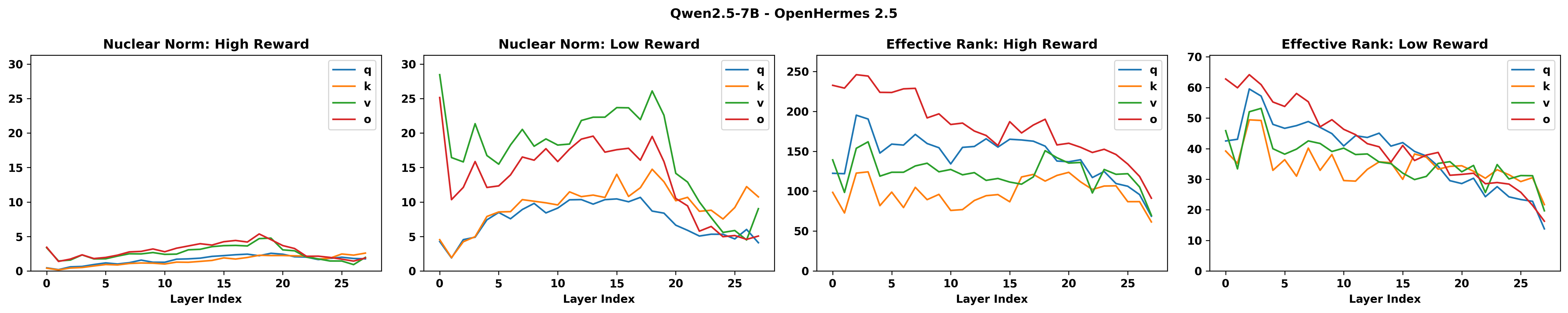}
    \caption{Qwen2.5 7B - OpenHermes with Reward Model Metric}
    \label{fig:qwen25_7b_her_reward}
\end{figure*}









\end{document}